\documentclass{article}

\usepackage{amsmath,amsfonts,bm}

\def\eqref#1{equation~\ref{#1}}

\def\1{\bm{1}}

\DeclareMathAlphabet{\mathsfit}{\encodingdefault}{\sfdefault}{m}{sl}
\SetMathAlphabet{\mathsfit}{bold}{\encodingdefault}{\sfdefault}{bx}{n}

\DeclareMathOperator*{\argmin}{arg\,min}

\usepackage{microtype}
\usepackage{graphicx}
\usepackage{subfigure}
\usepackage{booktabs} 

\usepackage{hyperref}
\usepackage{amsfonts}       
\usepackage{nicefrac}       
\usepackage{xcolor}         
\usepackage{subfigure}
\usepackage{wrapfig,lipsum,booktabs}
\usepackage{caption}

\usepackage[numbers]{natbib}

\usepackage[accepted]{icml2024}

\usepackage{amsmath}
\usepackage{amssymb}
\usepackage{mathtools}
\usepackage{amsthm}

\usepackage[capitalize,noabbrev]{cleveref}

\theoremstyle{plain}
\newtheorem{theorem}{Theorem}[section]

\theoremstyle{definition}

\theoremstyle{remark}

\renewcommand{\eqref}[1]{Eq.~(\textup{\ref{#1}})}

\usepackage[textsize=tiny]{todonotes}

\icmltitlerunning{Rethinking Guidance Information to Utilize Unlabeled Samples: A Label-Encoding Perspective}
\begin{document}

\twocolumn[
\icmltitle{Rethinking Guidance Information to Utilize Unlabeled Samples:\\A Label Encoding Perspective}

\icmlsetsymbol{equal}{*}

\begin{icmlauthorlist}
\icmlauthor{Yulong Zhang}{equal,sch,yyy}
\icmlauthor{Yuan Yao}{equal,comp}
\icmlauthor{Shuhao Chen}{sch}
\icmlauthor{Pengrong Jin}{sch}
\icmlauthor{Yu Zhang}{sch}
\icmlauthor{Jian Jin}{jin}
\icmlauthor{Jiangang Lu}{yyy}
\end{icmlauthorlist}

\icmlaffiliation{yyy}{College of Control Science and Engineering, Zhejiang University, Hangzhou, China}
\icmlaffiliation{comp}{Beijing Teleinfo Technology Company Ltd., China Academy of Information and Communications Technology, Beijing, China}
\icmlaffiliation{jin}{Research Institute of Industrial Internet of Things, China Academy of Information and Communications Technology, Beijing, China}
\icmlaffiliation{sch}{Department of Computer Science and Engineering, Southern University of Science and Technology, Shenzhen, China}

\icmlcorrespondingauthor{Yu Zhang}{yu.zhang.ust@gmail.com}
\icmlcorrespondingauthor{Jian Jin}{jin.jian@caict.ac.cn}
\icmlcorrespondingauthor{Jiangang Lu}{lujg@zju.edu.cn}

\icmlkeywords{Machine Learning, ICML}

\vskip 0.3in
]

\printAffiliationsAndNotice{\icmlEqualContribution}

\begin{abstract}
Empirical Risk Minimization (ERM) is fragile in scenarios with insufficient labeled samples. A vanilla extension of ERM to unlabeled samples is Entropy Minimization (EntMin), which employs the soft-labels of unlabeled samples to guide their learning. However, EntMin emphasizes prediction discriminability while neglecting prediction diversity. To alleviate this issue, in this paper, we rethink the guidance information to utilize unlabeled samples. By analyzing the learning objective of ERM, we find that the guidance information for labeled samples in a specific category is the corresponding \textit{label encoding}. Inspired by this finding, we propose a Label-Encoding Risk Minimization (LERM).
It first estimates the label encodings through prediction means of unlabeled samples and then aligns them with their corresponding ground-truth label encodings. As a result, the LERM ensures both prediction discriminability and diversity, and it can be integrated into existing methods as a plugin. Theoretically, we analyze the relationships between LERM and ERM as well as EntMin. Empirically, we verify the superiority of the LERM under several label insufficient scenarios.
The codes are available at 
\href{https://github.com/zhangyl660/LERM}{https://github.com/zhangyl660/LERM}.
\end{abstract}

\section{Introduction}

With abundant high-quality human-annotated samples, deep learning techniques have achieved remarkable advances in various applications
\cite{lecun2015deep,he2016deep,vaswani2017attention}.
One key principle behind their success is the Empirical Risk Minimization (ERM), which adopts the ground-truth labels of labeled samples to guide their learning. 
In practice, however, we often encounter some label insufficient scenarios \citep{cui2020towards}, where the labeled samples are limited or may even be absent altogether. For the former, we can utilize a large number of unlabeled samples to assist the learning of labeled samples, which falls within the scope of semi-supervised learning \citep{sohn2020fixmatch, zhang2021flexmatch, chen2022debiased}. On the contrary, for the latter, a popular solution is to borrow the knowledge from a related label-sufficient domain, \textit{i.e.}, source domain, for facilitating the learning of unlabeled samples, which pertains to the field of transfer learning \citep{pan2010survey, yang2020transfer}.
The commonality of those techniques is to fully utilize unlabeled samples for improving the generalization capability in scenarios with insufficient labeled samples. Such scenarios are frequently encountered in practical applications. 
As ERM heavily relies on the guidance of label information, it fails to fully exploit the potential of unlabeled samples. Accordingly, it does not achieve good performance in label insufficient scenarios.
Hence, this paper focuses on mining the potential of unlabeled samples to deal with label insufficient scenarios.

To achieve this, a simple and popular approach is the entropy minimization (EntMin) \citep{grandvalet2004semi}. It can be regarded as a direct extension of ERM to unlabeled samples. Specifically, it utilizes the soft-labels of unlabeled samples, which are assigned by a learning model during the training process, for guiding their learning. As a result, it pushes samples far from the decision boundary, thereby enhancing the prediction discriminability of unlabeled samples. 
However, one potential flaw of EntMin is that the soft-labels assigned by the learning model could be mainly from majority categories with large numbers of labeled samples, resulting in a decrease in prediction diversity \citep{cui2020towards} since the unlabeled samples tend to be biased toward the majority categories.
One reason for that lies in the absence of more appropriate guidance information for unlabeled samples. This leads us to ask a question: ``\textit{For unlabeled samples, is there more precise guidance information available?}''

To seek a potential solution to the above problem, we delve into the learning objective of the ERM.
Based on the ERM principle, we observe that all samples associated with a specific category need to be mapped to a label encoding, \textit{i.e.}, one-hot label encoding, corresponding to that category.
In other words, 
\textit{the guidance information used in the ERM for the labeled samples in a specific category is the corresponding label encoding}\footnote{Label encoding refers to one-hot label encoding by default.}. 
Moreover, under label insufficient scenarios studied in this paper, 
the label encodings of labeled samples remain consistent with those of unlabeled samples.
Accordingly, it is reasonable to apply label encodings as guidance information to supervise the learning of unlabeled samples.
Inspired by this finding, we propose the Label-Encoding Risk Minimization (LERM), a generalization of ERM, to handle unlabeled samples. Specifically, the proposed LERM first estimates the label encodings based on unlabeled samples by calculating their prediction means.
Then, the LERM minimizes the label-encoding risk, \textit{i.e.}, the divergence between the estimated and ground-truth label encodings.
Since those label encodings serve as accurate supervision information, in conjunction with many existing methods, the LERM can enhance their generalization capability under different label insufficient scenarios.

The contributions of this paper are highlighted as follows.
\begin{itemize}
\vspace{-2ex}
\setlength{\itemsep}{-1pt}
\item We find that the label encodings can serve as precise guidance information to supervise the learning of unlabeled samples. Also, we theoretically reveal that the prediction means of unlabeled samples can be used as estimations for label encodings.
\item The LERM is proposed to utilize unlabeled samples and it can be seamlessly integrated as a plugin into existing methods. Moreover, we provide a theoretical analysis to explore the relationships between LERM and ERM, as well as between LERM and EntMin.
\item Extensive experimental results are presented under several label insufficient scenarios, including semi-supervised learning (SSL), unsupervised domain adaptation (UDA), and semi-supervised heterogeneous domain adaptation (SHDA), which verify the effectiveness of the proposed LERM method.
\end{itemize}

\section{Related Work}
\label{gen_inst}

In this paper, we mainly focus on three typical tasks under label insufficient scenarios, \textit{i.e.}, Semi-Supervised Learning (SSL) \citep{sohn2020fixmatch, zhang2021flexmatch, chen2022debiased}, Unsupervised Domain Adaptation (UDA) \citep{ganin2016domain, long2018conditional, chen2024large, rangwani2022closer, zhang2023domain}, and Semi-supervised Heterogeneous Domain Adaptation (SHDA) \citep{yao2019heterogeneous, li2020simultaneous, gu2022keypoint, fang2022semi}.
Specifically, SSL leverages limited labeled samples and massive unlabeled samples to improve the generalization capability.
For example, FlexMatch \citep{zhang2021flexmatch} utilizes curriculum pseudo labeling for enhancing the performance of SSL, and DST \citep{chen2022debiased} mitigates the impact of incorrect pseudo-labels during the iterative self-training process. 
On the other hand, UDA aims to improve the learning of a target domain with unlabeled samples by harnessing the knowledge from a source domain with sufficient labeled samples. For instance, DANN \citep{ganin2016domain} and CDAN \citep{long2018conditional} bridge the source and target domains through adversarial training. AFN \citep{xu2019larger} progressively adapts the feature norms of the two domains to a broad range of values.
Recently, SDAT \citep{rangwani2022closer} enhances the stability of domain adversarial training and seeks a flat minimum.
Considering the heterogeneity of the features across domains, SHDA leverages a limited number of labeled samples from the target domain to improve the transfer performance. 
As an example, STN \citep{yao2019heterogeneous} adopts the soft-labels of unlabeled target samples to align the conditional distributions across domains. Another example is KPG \citep{gu2022keypoint}, which utilizes key samples to guide the matching process in optimal transport~\citep{villani2009optimal, wang2024optimal}.

Among those tasks, to boost performance, it is vital to utilize unlabeled samples. \mbox{EntMin} \citep{grandvalet2004semi} is commonly employed to enhance the prediction discriminability of unlabeled samples. For example, in SSL, \cite{berthelot2019mixmatch} utilizes EntMin to estimate low-entropy labels for data-augmented unlabeled samples. Moreover, \cite{yin2022fishermatch} adopts the entropy of the predicted distribution as a confidence measure to filter pseudo-labels effectively. In UDA, EntMin is utilized in \citep{long2016unsupervised, vu2019advent} to obtain a more reliable decision boundary in the unlabeled target domain. 
In addition, BNM \citep{cui2020towards} is proposed to maximize the nuclear norm of the prediction matrix derived from unlabeled samples, which ensures higher prediction discriminability and diversity. 
Similar to LERM, both EntMin and BNM can also be embedded into any existing SSL, UDA, and SHDA approaches. However, unlike them, the proposed LERM method adopts label encodings as the supervision information, which could achieve both the prediction discriminability and diversity.

\begin{figure*}[t]
\centering
\subfigure[ERM]{
    \includegraphics[width=0.39\textwidth]{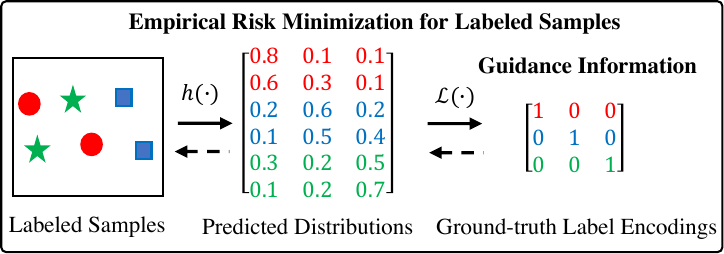}
    \label{fig:erm}
}
\hspace{-2ex}
\vspace{-1ex}
\subfigure[LERM]{
    \includegraphics[width=0.582\textwidth]{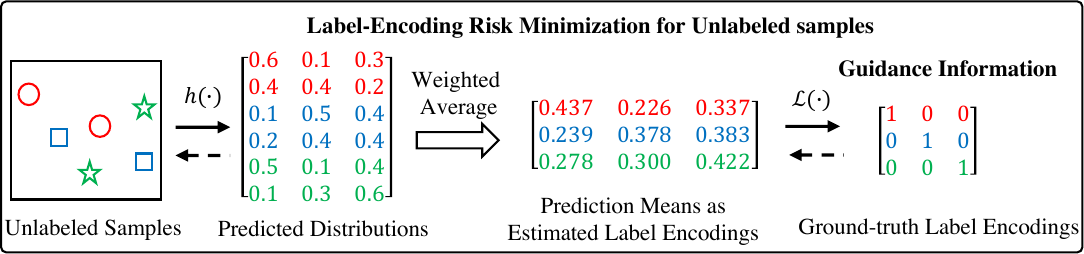}
    \label{fig:lrm}
}
\includegraphics[width=1.2\columnwidth]{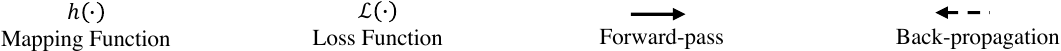}
\caption{Illustrations of the ERM and LERM. Here, different shapes denote distinct categories. In the ERM, we can observe that the six labeled samples are mapped into three label encodings associated with distinct categories. Also, the label encodings of labeled samples remain consistent with those of unlabeled samples. This inspires us to apply those label encodings as guidance information to supervise the learning of unlabeled samples. To this end, we propose the LERM. It first estimates the label encodings through prediction means for unlabeled samples and then aligns them with their corresponding ground-truth label encodings.}
\label{fig:illustration}
\vspace{-2ex}
\end{figure*}

\section{Preliminary}

The ERM provides learning guidance for labeled samples, which aims to choose an optimal function $h^* (\cdot)$ from a set of potential function sets $\mathcal{H}$. Here, $h(\cdot) \in \mathcal{H}$ characterizes the connection between a sample $\mathbf{x}$ and its corresponding label $\mathbf{y}$ with the joint distribution $P(\mathbf{x}, \mathbf{y})$.
To this end, we first need a non-negative real-valued loss function, represented as $\mathcal{L}(\cdot, \cdot)$. This function tells us how much it hurts to make the prediction $h(\mathbf{x})$ when the actual label is $\mathbf{y}$, for a labeled sample $(\mathbf{x}, \mathbf{y})\sim P(\mathbf{x}, \mathbf{y})$. 
Then, we can calculate the expectation of the loss function $\mathcal{L}(\cdot, \cdot)$ over the distribution $P(\mathbf{x}, \mathbf{y})$, \textit{i.e.}, the expected risk, by
\begin{equation}
\mathcal{R}_{exr} (h) = \mathbb{E}_{P} [\mathcal{L}(h(\mathbf{x}), \mathbf{y})] = \int \mathcal{L}(h(\mathbf{x}), \mathbf{y}) {\rm d} P(\mathbf{x}, \mathbf{y}).
\end{equation}

However, the distribution $P(\mathbf{x}, \mathbf{y})$ is unknown in most realistic scenarios. To handle this issue, we can calculate an estimation of the expected risk, \textit{i.e.}, the empirical risk, by averaging the loss function $\mathcal{L}(\cdot, \cdot)$ over a set of labeled samples $\mathcal{D}=\{(\mathbf{x}_i,\mathbf{y}_i)\}_{i=1}^n$, where $(\mathbf{x}_i,\mathbf{y}_i)\sim P(\mathbf{x}, \mathbf{y})$ for $i=1,...,n$. Accordingly, the empirical risk can be formulated as
\begin{equation}
\label{R_emr}
  \mathcal{R}_{emr}(h) = \frac{1}{n}\sum_{i=1}^{n}\mathcal{L}(h(\mathbf{x}_i), \mathbf{y}_i).
\end{equation}
Finally, we pick an optimal function $h^*(\cdot)$ that minimizes \eqref{R_emr}, which is known as the ERM:
\begin{equation}
h^*=\argmin_{h}\mathcal{R}_{emr}(h) =\argmin_{h} \frac{1}{n} \sum_{i = 1}^n \mathcal{L}(h(\mathbf{x}_i), \mathbf{y}_i).
\end{equation}

\section{Methodology}

In this section, we introduce the proposed LERM principle.

\subsection{A Motivating Example}

As aforementioned, in label insufficient scenarios, identifying accurate guidance for training unlabeled samples is critical. Analyzing the ERM's learning objective, exemplified in \cref{fig:erm}, reveals that six labeled samples across three categories are mapped to three distinct label encodings: $[1,0,0]$, $[0,1,0]$, and $[0,0,1]$. That is, in the ERM, the guidance information of the labeled samples in a specific category is the corresponding label encoding. Since in this paper we focus on the label insufficient scenarios where the labeled and unlabeled samples share the same categories, the label encodings remain consistent for both labeled and unlabeled samples. As depicted in \cref{fig:lrm}, there are a total of six unlabeled samples distributed across three different categories. Despite the absence of their ground-truth labels during training, the label encodings corresponding to these categories remain $[1,0,0]$, $[0,1,0]$, and $[0,0,1]$. This inspires us to use these label encodings as precise guidance information to guide the learning of unlabeled samples.

Unfortunately, label encodings for each unlabeled sample are unattainable, preventing their direct use in supervision like labeled samples. However, the one-to-one relationship between label encoding and category provides a solution: \textit{Estimating label encodings across all categories using unlabeled samples}.
Since each label encoding uniquely aligns with a specific category, the predicted category distribution by a learning model for each unlabeled sample attracts our attention. The predicted category distribution shows the probability of each sample belonging to different categories.
Thus, for a given category, we commence by obtaining the probabilities that all unlabeled samples belong to that category. Then, we calculate a weighted average over the predicted category distributions of all unlabeled samples, \textit{a.k.a.} \textit{prediction mean}, using those probabilities as weights. Accordingly, the prediction mean can be regarded as an estimation of the label encoding associated with the corresponding category (The mathematical principle behind this perspective will be explained later). As a result, we first estimate the label encodings through the prediction means and then minimize the divergence between them and their corresponding ground-truth label encodings. We refer to that divergence as \textit{label-encoding risk}, which is the core of the LERM. The key rationale of the LERM is illustrated in \cref{fig:lrm}. Next, we elaborate on the LERM principle.

\subsection{LERM}

We first introduce how to calculate the prediction means based on unlabeled samples. Let $f(\cdot)$ be a classifier and $g(\cdot)$ be a feature extractor in the form of deep neural networks. Given $n_u$ unlabeled samples $\{\mathbf{x}_i^u\}_{i=1}^{n_u}$,
we calculate the predicted category distribution of $\mathbf{x}_i^u$ as $\widetilde{\mathbf{y}}_{i}^u = f(g(\mathbf{x}_i^u)) \in \mathbb{R}^{C}$, 
where its $c$-th element $\widetilde{y}_{i,c}^u$ denotes the probability that $\mathbf{x}_{i}^u$ belongs to category $c$. Accordingly, the prediction mean for category $c$ is defined as
\begin{equation}
\label{m_c^u}
\mathbf{m}_c^u = \frac{1}{ \sum_{i = 1}^{n_u} {\widetilde{y}_{i,c}^{u}} } \big( \sum_{i = 1}^{n_u} {\widetilde{y}_{i,c}^{u}} \widetilde{\mathbf{y}}_{i}^u \big).
\end{equation}

For all the prediction means $\{\mathbf{m}_c^u\}_{c=1}^C$, their properties are summarized in \cref{theorem_m_property}, which explains why the prediction mean $\mathbf{m}_c^u$ could be treated as an estimation of the label encoding associated with category $c$.

\begin{theorem}
\label{theorem_m_property}

$\mathbf{m}_c^u$ satisfies the following properties:

(1) $\mathbf{1}^T\mathbf{m}_c^u=1$, where $\mathbf{1} \in \mathbb{R}^C$ denotes an all-ones vector.

(2) $0 \le m_{c,j}^u\le 1$, $\forall j \in \{1,\dots,C\}$, where $m_{c,j}^u$ denotes the $j$-th element of $\mathbf{m}_c^u$.

(3) If $\widetilde{\mathbf{y}}_{i}^u$ equals the label encoding of the ground-truth label of sample $\mathbf{x}_i^u$ for each $i \in \{1, \dots, n_u\}$, then $\mathbf{m}_c^u$ equals $\mathbf{e}_c$. Here, $\mathbf{e}_c$ denotes the one-hot label encoding of category $c$ with its $c$-th element as 1 and other elements as 0.

(4) If $\mathbf{m}_c^u$ equals $\mathbf{e}_c$ for some $c \in \{1, \cdots, C\}$, then for any $i \in \{1, \cdots, n_u\}$, $\widetilde{\mathbf{y}}_{i}^u$ either equals $\mathbf{e}_c$ or satisfies the condition that $\widetilde{y}_{i,c}^u = 0$, $0 \leq \widetilde{y}_{i,k}^{u} \leq  1$, $\forall k \neq c$. 

(5) If $\mathbf{m}_c^u$ equals $\mathbf{e}_c$ for any $c \in \{1, \cdots, C\}$, then for any $i \in \{1, \cdots, n_u\}$, $\widetilde{\mathbf{y}}_{i}^u$ is a one-hot vector with only one element equal to 1 and other elements being 0.
\end{theorem}

The proof of \cref{theorem_m_property} can be found in \cref{prove_m_property}.
Based on property (3) in \cref{theorem_m_property}, we find that when $\widetilde{\mathbf{y}}_{i}^u$ approaches the ground-truth label encoding of sample $\mathbf{x}_i^u$, $\mathbf{m}_c^u$ tends to approach the ground-truth label encoding associated with category $c$, \textit{a.k.a.} $\mathbf{e}_c$. Accordingly, $\mathbf{m}_c^u$ could be regarded as an estimation for $\mathbf{e}_c$. 

Building upon the above theoretical perspectives, we formulate the label-encoding risk as
\begin{equation}
\label{Rler}
\mathcal{R}_{ler}(f, g)=\frac{1}{C}\sum_{c=1}^{C}\mathcal{L}(\mathbf{m}_c^u, \mathbf{e}_c),
\end{equation}
where $\mathcal{L} (\cdot, \cdot)$ denotes a loss function to measure the divergence between the two input arguments. 
By minimizing the label-encoding risk, we could make the estimated label encodings close to the corresponding ground-truth ones. As a result, 
the learning of unlabeled samples can be effectively guided by ground-truth label encodings.
We refer to such risk minimization principle as the LERM, which can be utilized as a plugin for existing methods to handle label insufficient scenarios. 

\subsection{Discussion with Existing Studies} 

\subsubsection{Connection between LERM and ERM}
\label{sec:connect_LERM_ERM}

ERM is susceptible to overfitting when dealing with extremely label-scarce scenarios because it does not utilize unlabeled samples. Instead, LERM, specifically designed for handling unlabeled samples, considers both prediction discriminability and diversity. As a result, LERM can be regarded as a regularization term for model parameters, helping mitigate the potential risk of overfitting in existing methods that are integrated with LERM. Furthermore, since LERM draws inspiration from ERM, we next analyze their relationship within the supervised learning framework.

Under the above framework, the category information of each labeled sample is known and accurate. Hence, we can utilize that category information to calculate the prediction means of labeled samples, thereby eliminating interference from samples in other categories.
Concretely, for a given category, we first acquire all labeled samples associated with that category. Then, we calculate an average over the predicted category distributions of those labeled samples. 
Accordingly, the prediction mean of labeled samples belonging to category $c$ is defined by
\begin{equation}
\label{m_c^l}
\mathbf{m}_c^l = \frac{1}{n_c^l} \sum_{i = 1}^{n_c^l} f (g(\mathbf{x}_{i}^{l, c})) = \frac{1}{n_c^l} \sum_{i = 1}^{n_c^l} \widetilde{\mathbf{y}}_{i}^{l, c},
\end{equation}
where $\mathbf{x}_{i}^{l, c}$ is the $i$-th labeled sample belonging to category $c$, $\widetilde{\mathbf{y}}_{i}^{l, c}$ is the predicted category distribution of $\mathbf{x}_{i}^{l, c}$, and $n_c^l$ is the number of labeled samples associated with category $c$. Similar to \eqref{Rler}, the label-encoding risk for labeled samples can be formulated as
\begin{equation}
\label{Rler_L}
\mathcal{R}_{ler}^l (f, g)=\frac{1}{C}\sum_{c=1}^{C}\mathcal{L}(\mathbf{m}_c^l, \mathbf{e}_c).
\end{equation}

Based on the above definitions, we reveal the theoretical relationship between LERM and ERM in \cref{thm2}.
\begin{theorem}
\label{thm2}
Under the setting of supervised learning, if both the label-encoding and empirical risks utilize the same loss function which is convex w.r.t. the first input argument and $\frac{1}{n_l} \sum_{c=1}^{C} n_c^l \mathcal{L} (\mathbf{m}_{c}^{l}, \mathbf{e}_c) \ge \frac{1}{C}\sum_{c=1}^{C} \mathcal{L} \left( \mathbf{m}_{c}^{l}, \mathbf{e}_c \right)$ holds, then the label-encoding risk is upper-bounded by the empirical risk.
\end{theorem}

The proof of \cref{thm2} is offered in \cref{appendixB}.

\subsubsection{Connection between LERM and EntMin}
\label{sec:comp_EntMin}

According to properties (4) and (5) in Theorem \ref{theorem_m_property}, as the label-encoding risk decreases, $\widetilde{\mathbf{y}}_{i}^u$ tends to be a one-hot vector (please refer to \cref{analysis} for empirical evidence).
EntMin aims to achieve a similar behavior by minimizing the entropy over $\widetilde{\mathbf{y}}_{i}^u$, while LERM minimizes the label-encoding risk.
Hence, on the one hand, both LERM and EntMin can enhance the prediction discriminability by pushing unlabeled samples away from the decision boundary but in different ways. 

On the other hand, according to the definition of the label-encoding risk for unlabeled samples in \eqref{R_emr}, we can see that each category contributes one loss with a weight of $\frac{1}{C}$ to the label-encoding risk.
Accordingly, \textit{LERM is category-specific and optimizes all the categories with the same weight, thereby mitigating the dominance of majority categories} (please refer to \cref{AppendixPDA} for empirical evidence). Recall that the entropy for unlabeled samples is formulated as
\begin{align}
\label{EntMin}
\begin{scriptsize}
\mathcal{R}_{ent}(f,g) \!=\! -\frac{1}{n_u} \!\sum_{i=1}^{n_u} \!\sum_{c=1}^{C} \widetilde{y}_{i,c}^u \!\ln \widetilde{y}_{i,c}^u.
\end{scriptsize}
\end{align}
Here, we can observe that \textit{EntMin is sample-specific and it is prone to be dominated by majority categories with a large number of unlabeled samples}.
Accordingly, this may lead to misclassification of unlabeled samples into those majority categories, resulting in limited prediction diversity.

Moreover, in \cref{thm3} we reveal the theoretical relationship between LERM and EntMin.
\begin{theorem}
\label{thm3}
If the label-encoding risk utilizes the cross-entropy loss function, \textit{i.e.}, $\mathcal{L} (\mathbf{m}_c^u, \mathbf{e}_c) = - \mathbf{e}_c^\top \ln (\mathbf{m}_c^u)$, and the inequality $\frac{1}{n_u}\sum_{c=1}^{C} (\sum_{j=1}^{n_u}\widetilde{y}_{j,c}^u) \mathcal{L} (\mathbf{m}_c^u, \mathbf{e}_c) \ge \frac{1}{C}\sum_{c=1}^{C} \mathcal{L} (\mathbf{m}_c^u, \mathbf{e}_c)$ holds, then the label-encoding risk is upper-bounded by the entropy regularization used in the EntMin.
\end{theorem}
The proof of \cref{thm3} is offered in \cref{appendixC}. 

\subsection{Application to Label Insufficient Scenarios}

In this section, we detail how to apply LERM to three label insufficient scenarios, \textit{i.e.}, SSL, UDA, and SHDA. 

\subsubsection{SSL}

Under the SSL setting, we have $n_l$ labeled samples $\{(\mathbf{x}_i^l,\mathbf{y}_i^l)\}_{i=1}^{n_l}$, where $\mathbf{y}_i^l$ is the label encoding of $\mathbf{x}_i^l$.
Also, we have $n_u$ unlabeled samples $\{\mathbf{x}_i^u\}_{i=1}^{n_u}$.
Here, we have $n_l \ll n_u$. The objective of SSL is to learn a good model for predicting labels of $\{\mathbf{x}_i^u\}_{i=1}^{n_u}$. To this end, we can plug the LERM into existing SSL approaches.
In addition, following the setting of supervised pretraining in \citep{chen2022debiased}, we augment the labeled and unlabeled samples by leveraging a weak augmentation function, \textit{i.e.}, $\psi(\cdot)$, and a strong augmentation, \textit{i.e.}, $\Psi(\cdot)$. 
Thus, by utilizing LERM, the overall objective function in SSL is formulated as
\begin{align}
\label{loss_ssl}
\min_{{f}, {g}} & \frac{1}{n_l}\!\! \sum_{i = 1}^{n_l} \!\mathcal{L}_{ce}\! \Big[ \!{f}\! \big( {g}(\psi(\mathbf{x}_i^l)\!)\! \big), \mathbf{y}_i^l \Big] 
\!\!+\!\! \frac{\mu}{n_l} \!\!\sum_{i = 1}^{n_l}\! \mathcal{L}_{ce}\! \Big[ {f} \!\big({g}(\Psi(\mathbf{x}_i^l)\!)\! \big), \mathbf{y}_i^l \Big] 
\notag\\
& + \alpha \mathcal{L}_{ssl}+ \frac{\lambda}{C} \sum_{c = 1}^C \Big[ \mathcal{L} ({\mathbf{w}}_c^u, \mathbf{e}_c) + {\mu} \mathcal{L} ({\mathbf{s}}_c^u, \mathbf{e}_c) \Big],
\end{align}
where $\mathcal{L}_{ce}(\cdot, \cdot)$ denotes the cross-entropy loss, 
$\mathcal{L}_{ssl}$ denotes the semi-supervised learning loss of an existing SSL method if any, ${\mu}$, $\alpha$, and ${\lambda}$ are three trade-off hyperparameters, and ${\mathbf{w}}_c^u$ and ${\mathbf{s}}_c^u$ denote the prediction means under the weak and strong augmentations, respectively. Hence, ${\mathbf{w}}_c^u$ and ${\mathbf{s}}_c^u$ can be computed according to \eqref{m_c^u} by replacing $\mathbf{x}_i^u$ with $\psi(\mathbf{x}_i^u)$ and $\Psi(\mathbf{x}_i^u)$, respectively. 
When $\alpha$ is set to zero, the problem in (\ref{loss_ssl}) degenerates into minimizing a combination of the ERM and LERM, and the same situation also occurs in the following two tasks.

\subsubsection{UDA}

For a UDA task, we are offered $n_s$ labeled source samples $\{(\mathbf{x}_i^s,\mathbf{y}_i^s)\}_{i=1}^{n_s}$ and $n_t$ unlabeled target samples $\{\mathbf{x}_i^u\}_{i=1}^{n_t}$.
The goal is to learn a high-quality model for categorizing $\{\mathbf{x}_i^u\}_{i=1}^{n_t}$. To achieve this, we incorporate the LERM into existing UDA methods. Accordingly, we can formulate the objective function as
\begin{align}
\label{loss_uda}
\tiny
\min_{{f}, {g}} & \frac{1}{n_s} \!\sum_{i = 1}^{n_s} \!\mathcal{L}_{ce} \!\Big[ {f} ({g}({\mathbf{x}}_i^s)), {\mathbf{y}}_i^s \Big]
\!+\! \alpha \mathcal{L}_{uda} \!+\! \frac{{\lambda}}{C} \!\sum_{c = 1}^C \! \mathcal{L} ({\mathbf{m}}_c^u,\mathbf{e}_c),
\end{align}
where $\mathcal{L}_{uda}$ denotes the domain adaptation loss of an existing UDA method, $\alpha$ and $\lambda$ are two trade-off hyperparameters, and the prediction mean ${\mathbf{m}}_c^u$ is obtained by \eqref{m_c^u} based on unlabeled target samples.

\begin{table*}[!t]\footnotesize
\centering
\caption{Accuracy (\%) comparison on the CIFAR-10, CIFAR-100, DTD, and ImageNet-1K datasets under the SSL setting. The best performance of each task is marked in bold and the best performance in each comparison group is underlined.}
\label{SSL}
\begin{tabular}{c|cc|c|cc|c|cc|c|cc}
\toprule Dataset               & \multicolumn{3}{c|}{CIFAR-10}    & \multicolumn{3}{c|}{CIFAR-100}   & \multicolumn{3}{c|}{DTD}   & \multicolumn{2}{c}{ImageNet-1K}    \\
\cmidrule(r){2-12}  
\# Label per category & \multicolumn{2}{c|}{1} & 4       & \multicolumn{2}{c|}{1} & 4       & \multicolumn{2}{c|}{1} & 4   & \multicolumn{2}{c}{100}  \\
& Top-1      & Top-5    & Top-1   & Top-1     & Top-5     & Top-1   & Top-1     & Top-5     & Top-1 & Top-1& Top-5\\
\midrule
ERM    & 32.24  & 78.16  & 57.04   & 23.58     & 47.51     & 47.18   & 31.22     & 58.99     & 50.66  & 44.98    & 69.00 \\
ERM + EntMin    & 28.17      & 71.05    & 59.62   & 15.32     & 43.95     & 45.40   & 21.55     & 51.65     & 50.96 & 49.26    & 72.60  \\
ERM + BNM      & 27.02      & 70.37    & 52.46   & 21.79     & 47.72     & 58.90   & 28.55     & 54.61     & 48.26  & 49.81   & 72.73\\
ERM + LERM & \underline{38.22}   & \underline{80.82}  & \underline{75.57}  & \underline{30.15}  & \underline{61.33} &\underline{60.19}  &  \underline{34.84}   &  \underline{63.51} & \underline{53.14}  & \underline{50.83}  & \underline{74.11} \\
\midrule
FlexMatch    & 40.86      & 84.75    & 86.66   & 16.49     & 42.40     & 65.11   & 33.39     & 58.48 & 54.96 & 50.34  & 75.02\\
FlexMatch + EntMin      & 43.79      & 87.69    & 86.56   & 13.00 & 42.83 & 67.32   & 32.20     & 58.49     & 54.91 & 53.26  & 76.99\\
FlexMatch + BNM         & 41.95      & 78.73    & 86.57   & 15.04 & 43.54 & 64.46   & 31.31     & 57.31     & 55.04 & 55.12  & 78.62\\
FlexMatch + LERM  & \underline{53.69} & \underline{89.18} & \underline{88.28} & \underline{19.50} & \underline{46.00} & \underline{\textbf{69.65}} & \underline{34.42}  &  \underline{58.51} & \underline{55.11} & \underline{\textbf{56.69}}  & \underline{\textbf{79.79}}\\
\midrule
DST     & 51.11      & 91.76    & 88.05   & 32.92     & 64.65     & 66.80   & 34.88     & 61.99     & 56.40 & 50.34 & 75.94\\
DST + EntMin            & 45.46      & 92.41    & 87.85   & 25.48     & 60.92     & 66.79   & 32.32     & 62.27     & 56.13 & 53.82  & 76.28\\
DST + BNM               & 55.03      & 91.75    & 88.49   & 32.15     & 65.16     & 67.27   & 36.08     & 64.06     & 56.51 & 54.28 & 76.56\\
DST + LERM & \underline{\textbf{62.04}} & \underline{\textbf{93.09}}  & \underline{\textbf{89.71}}  & \underline{\textbf{43.78}}   & \underline{\textbf{70.37}}     & \underline{68.65}  & \underline{\textbf{38.19}}  &  \underline{\textbf{67.39}}  & \underline{\textbf{57.45}} & \underline{54.60}  & \underline{76.87}\\
\bottomrule
\end{tabular}
\vspace{-3ex}
\end{table*}

\subsubsection{SHDA}

In the SHDA problem, we are given $n_s$ labeled source samples $\{(\mathbf{x}_i^s,\mathbf{y}_i^s)\}_{i=1}^{n_s}$, $n_l$ labeled target samples $\{(\mathbf{x}_i^l,\mathbf{y}_i^l)\}_{i=1}^{n_l}$, and $n_t$ unlabeled target samples $\{\mathbf{x}_i^u\}_{i=1}^{n_t}$.
Here, we have $n_s \gg n_l$ and $n_t \gg n_l$. 
The goal is to learn a model for classifying $\{\mathbf{x}_i^u\}_{i=1}^{n_t}$. As the heterogeneity of the features across domains, we let $g_s(\cdot)$ and $g_t(\cdot)$ represent the feature extractors in the source and target domains, respectively. For the above purpose, we embed the LERM within established SHDA models. 
In addition, following the setting in \citep{yao2019heterogeneous}, we utilize an additional regulation term to prevent overfitting. 
As a result, the entire objective function is formulated as
\begin{align}
\label{loss_shda}
\min_{{f}, g_s, g_t} &\frac{1}{n_s}\!\sum_{i = 1}^{n_s}\!\mathcal{L}_{ce}\! \Big[\! f(g_s({\mathbf{x}}_i^s)), \mathbf{y}_i^s \Big] \!+\! \frac{1}{n_l}\!\sum_{i = 1}^{n_l}\!\mathcal{L}_{ce}\! \Big[ \!f(g_t({\mathbf{x}}_i^l)), \mathbf{y}_i^l \Big] 
\nonumber \\
& \!\!\!+\! \alpha \mathcal{L}_{shda} \!\!+\!\! \frac{{\lambda}}{C}\!\sum_{c = 1}^C\!\mathcal{L}({\dot{\mathbf{m}}}_c^u, \mathbf{e}_c) 
 \!\!+\!\! {\tau} (\| {f} \|^2 \!\!+\!\! \| g_s \|^2 \!\!+\!\! \| g_t \|^2),
\end{align}
where $\mathcal{L}_{shda}$ is the domain adaptation loss of an existing SHDA method, ${\alpha}$, ${\lambda}$, and $\tau$ act as three trade-off hyperparameters, and the prediction mean ${\dot{\mathbf{m}}}_c^u$ is computed similar to \eqref{m_c^u} by replacing $g(\cdot)$ with $g_t(\cdot)$.

\begin{table*}[!t]\small
\centering
\caption{Accuracy (\%) comparison on the Office-Home dataset under the UDA setting. The best performance of each task is marked in bold and the best performance in each comparison group is underlined.}
\label{officehome}
\setlength{\tabcolsep}{0.5mm}{
\begin{tabular}{*{14}{c}}
\toprule Method & Ar$\rightarrow$Cl   & Ar$\rightarrow$Pr  & Ar$\rightarrow$Rw & Cl$\rightarrow$Ar & Cl$\rightarrow$Pr & Cl$\rightarrow$Rw & Pr$\rightarrow$Ar & Pr$\rightarrow$Cl & Pr$\rightarrow$Rw & Rw$\rightarrow$Ar & Rw$\rightarrow$Cl & Rw$\rightarrow$Pr & Average \\
\midrule
ERM & 44.00 & 67.16 & 74.19 & 52.98 & 61.65 & 64.29 & 52.12 & 39.10 & 73.01 & 64.33 & 43.73 & 75.29 & 59.32 \\
DANN & 52.84 & 62.90 & 73.46 & 56.26 & 67.42 & 68.01 & 58.40 & 54.41 & 78.92 & 70.65 & \underline{60.31} & 80.79 & 65.36  \\
AFN  & 52.68 & 72.27 & 76.96 & \underline{65.13} & 71.13 & 72.78 & \underline{63.93} & 51.33 & 77.81 & \underline{72.12} & 57.52 & 81.98 & 67.97 \\
ERM + EntMin  & 46.83 & 67.28 & 77.24 & 62.23 & 70.26 & 71.69 & 59.22 & 47.06 & 79.34 & 70.92 & 54.47 & 82.43 & 65.75 \\
ERM + BNM  & 54.78 & \underline{74.86} & \underline{79.51} & 63.04 & 72.00 & \underline{74.99} & 61.41 & 52.21 & \underline{80.27} & 71.39 & 57.33 & 82.77 & 68.71 \\
ERM + LERM & \underline{55.56} & 74.61	&79.48&	64.48&	\underline{73.71}&	74.45&	62.55&	\underline{52.23} &	79.96 &	71.82 &	57.98 &	\underline{83.31} &	\underline{69.18} \\
\midrule
CDAN  & 55.18 & 72.63 & 78.01 & 62.01 & 72.46 & 73.11 & 62.68 & 53.99 & 79.65 & 72.83 & 58.23 & 83.60 & 68.70 \\
CDAN + EntMin & 54.82 & 72.43 & 78.90 & 63.03 & 72.51 & 72.87 & 62.23 & 53.53 & 80.04 & 72.42 & 58.14 & 83.73 & 68.72 \\
CDAN + BNM & 56.00 & 74.00 & 78.94 & 63.59 & 73.31 & 73.79 & 62.53 & 53.91 & 81.05 & \underline{73.03} & 59.08 & 83.55 & 69.40 \\
CDAN + LERM & \underline{56.20}&	\underline{74.57} &	\underline{79.44}&	\underline{64.15}&	\underline{75.24} &	\underline{74.98} &	\underline{63.04} &	\underline{56.20}&	\underline{81.32} &	72.64&	\underline{59.36} &	\underline{84.41}	&\underline{70.13} \\
\midrule
SDAT  & 57.66 & 77.06 & 81.30 & 66.07 & 76.14 & 75.91 & 63.23 & 55.92 & 81.85 & 75.87 & 62.36 & 85.41 & 71.57 \\
SDAT + EntMin & 56.97 & 77.74 & 81.33 & 65.90 & 75.83 & 75.99 & 63.58 & 55.36 & 82.30 & 75.09 & 62.01 & 85.32 & 71.45 \\
SDAT + BNM & 57.59 & 76.95 & 80.84 & 66.13 & 75.33 & 75.65 & 65.49 & 56.07 & 81.87 & 75.38 & 62.47 & 85.51 & 71.61 \\
SDAT + LERM & \underline{\textbf{58.35}}& \underline{\textbf{78.01}} & \underline{\textbf{82.01}} & \underline{\textbf{67.37}}& \underline{\textbf{77.77}}	& \underline{\textbf{77.07}} &  \underline{\textbf{66.54}} & \underline{\textbf{56.08}} & \underline{\textbf{82.65}} & \underline{\textbf{75.90}}& 	\underline{\textbf{64.17}}& 	\underline{\textbf{85.97}}& 	\underline{\textbf{72.66}} \\
\bottomrule
\vspace{-5ex}
\end{tabular}}
\end{table*}

\section{Experiments}
\label{Experiments}
We assess the performance of the LERM on three typical label insufficient scenarios, including SSL, UDA, and SHDA. The loss function defined in \eqref{Rler} adopts the $\ell_1$ distance, \textit{i.e.}, $\mathcal{L}(\mathbf{m}_c^u, \mathbf{e}_c)=\|\mathbf{m}_c^u-\mathbf{e}_c\|_1$, where $\|\cdot\|_1$ denotes the $\ell_1$ norm of a vector. For comparisons among different loss functions, please refer to \cref{appendix:loss}. 

\subsection{Results}

\textbf{Evaluation on SSL Tasks}. We evaluate the LERM on four SSL benchmark datasets, including 
CIFAR-10 \citep{krizhevsky2009learning}, CIFAR-100 \citep{krizhevsky2009learning}, DTD \citep{cimpoi2014describing}, and ImageNet-1K \citep{deng2009imagenet}.
We conduct experiments on the SSL tasks with limited labeled samples.
In the comparison experiments, we combine the LERM with ERM, and state-of-the-art SSL approaches such as FlexMatch \citep{zhang2021flexmatch} and DST \citep{chen2022debiased} to tackle all the above tasks. 
Also, we realize the EntMin and BNM in the same way as the LERM, and report the average classification accuracy of each method in three randomized trials. More experimental details can be found in the \cref{a_SSL}.

According to the results shown in Table \ref{SSL}, we can see that on the first three datasets (\textit{i.e.}, CIFAR-10, CIFAR-100, and DTD), the LERM yields significant performance improvements under all the settings.
Specifically, for the SSL tasks with four labeled samples per category, the LERM achieves accuracy improvements of 18.53\%, 13.01\%, and 2.48\% over the ERM method on the CIFAR-10, CIFAR-100, and DTD datasets, respectively, and it also performs better than EntMin and BNM.
Moreover, when combined with state-of-the-art SSL methods, the LERM could further improve the performance. 
For instance, LERM brings performance improvements of 4.54\% and 1.85\% over FlexMatch~\citep{zhang2021flexmatch} and DST~\citep{chen2022debiased} on the CIFAR-100 dataset, respectively.
Those results highlight the potential benefits of combining LERM with existing SSL approaches.
For the more challenging case with one labeled sample per category, both the EntMin and BNM exhibit varying degrees of performance degradation when compared with ERM.
On the contrary, the LERM shows consistent performance improvements on three benchmark datasets, which demonstrates its effectiveness in scenarios with extremely limited labeled samples. 
Moreover, the LERM further enhances the performance of FlexMatch and DST, and surpasses EntMin and BNM with a large margin in terms of top-1 accuracy.
Specifically, with only one labeled sample per category, LERM achieves a top-1 performance improvement of 12.83\% over the FlexMatch method on the CIFAR-10 dataset, demonstrating its superiority in better utilizing unlabeled samples.
We further validate the effectiveness of LERM on the ImageNet-1K dataset, offering a more realistic and intricate setting for evaluation. The experimental results of 100 labels per class show that LERM still achieves better performance than both EntMin and BNM.
In summary, LERM has shown promising results in improving the performance of existing SSL methods and outperforming both EntMin and BNM.

\begin{table}[!t]
\vspace{-1ex} 
\footnotesize
\centering
\caption{Accuracy (\%) comparison on the Office-31 dataset under the UDA setting. The best performance of each task is marked in bold and the best performance in each comparison group is underlined.}
\label{office31}
\setlength{\tabcolsep}{1.5mm}{
\resizebox{\columnwidth}{!}{
\begin{tabular}{ccccccc @{\hskip 0.1in} c}
\toprule Method & A$\rightarrow$D        & A$\rightarrow$W  & D$\rightarrow$W        & W$\rightarrow$D    & D$\rightarrow$A    & W$\rightarrow$A        & Average \\
\midrule
ERM  & 81.15 & 77.00 & 96.60 & 99.00 & 63.98 & 64.01 & 80.29 \\
DANN  & 83.62  & 89.30 & 97.81 & \underline{\textbf{100.00}} & 72.01 & 74.11 & 86.14 \\
AFN  & \underline{95.29} & 91.18 & 98.73 & \underline{\textbf{100.00}} & 72.13 & 70.60 & 87.99 \\
ERM + EntMin  & 87.42 & 88.01 & 98.49 & 99.93 & 68.04 & 61.83 & 83.95 \\
ERM + BNM  & 89.36 & 91.36 & 98.62 & 99.93 & 70.76 & 71.29 & 86.89 \\
ERM + LERM & 92.37 & \underline{92.96} & \underline{98.74} & \underline{\textbf{100.00}} & \underline{72.45} & \underline{72.81} & \underline{88.22} \\
\midrule 
CDAN  & 92.77 & 92.37 & 98.79 & \underline{\textbf{100.00}} & 72.46 & 70.31 & 87.78 \\
CDAN + EntMin & 92.64 & 91.49 & 98.87 & \underline{\textbf{100.00}} & 71.87 & 71.80 & 87.78 \\
CDAN + BNM & 92.17 & 92.87 & 99.20 & \underline{\textbf{100.00}} & 73.53 & 73.15 & 88.49 \\
CDAN + LERM & \underline{93.78} & \underline{\textbf{93.33}} & \underline{\textbf{99.25}} & \underline{\textbf{100.00}} & \underline{73.59} & \underline{74.30} & \underline{89.04} \\
\midrule
SDAT  & 94.99 & 89.77 & \underline{99.04} & \underline{\textbf{100.00}} & 77.04 & 72.73 & 88.93 \\
SDAT + EntMin & 95.58 & 93.04 & 98.74 & \underline{\textbf{100.00}} & 77.50 & 72.41 & 89.54 \\
SDAT + BNM & 95.58 & 92.91 & 98.66 & \underline{\textbf{100.00}} & 77.61 & 74.97 & 89.96 \\
SDAT + LERM & \underline{\textbf{96.79}} & \underline{93.21} & 98.87 & \underline{\textbf{100.00}} & \underline{\textbf{78.06}} & \underline{\textbf{75.26}}	&\underline{\textbf{90.37}} \\
\bottomrule
\vspace{-6ex}
\end{tabular}}}
\end{table}

\begin{table*}[!htbp]\small
\centering
\caption{Accuracy (\%) comparison on the VisDA dataset under the UDA setting. 
The best performance of each task is marked in bold and the best performance in each comparison group is underlined.
}
\label{tab:visda}
\setlength{\tabcolsep}{0.5mm}{
\begin{tabular}{*{14}{c}}
\toprule Method & aeroplane & bicycle & bus & car & horse & knife &	motor &	person & plant &	skate &	train &	truck & Average \\
\midrule
ERM  & 76.27 &	22.59 &	54.38 &	75.18 &	76.07 &	12.99 &	84.93 &	19.76&	79.34&	29.13 &	79.61&	5.38 &	51.30 \\
DANN  & \underline{94.75} & 73.47 & 83.46 & 47.91&	87.00 &	\underline{88.30} &	88.47 &	\underline{77.18} &	88.16&	\underline{90.05} &	87.21&	42.26&	79.02  \\
AFN  & 93.13 & 54.76 &	81.03 &	69.74 &	92.36 &	75.88 & 92.11 &	73.83&	93.16 &	55.55 &	\textbf{\underline{90.48}} &	23.63&	74.64 \\
ERM + EntMin  & 92.33 & 13.18 & \textbf{\underline{84.88}} & \textbf{\underline{80.77}} & 91.25 &	65.53 &	\textbf{\underline{94.81}} &	39.08&	92.75&	16.41&	85.49&	1.60& 63.17 \\
ERM + BNM  & 94.19  & \underline{81.56} & 77.11 &	66.58&	\underline{93.24}&	82.10&	86.27& 76.74 &	\underline{93.37} &	57.08 &	89.07 &	49.89&	78.93 \\
ERM + LERM & 93.72 & 80.09 & 75.14 & 72.02 &	92.35 &	86.70 &	90.25 &	76.70&	90.81 &	73.26 &	87.61&	\underline{49.96} & \underline{80.72} \\
\midrule
CDAN  & \underline{95.10}&	75.78&	\underline{82.30}&	57.52&	90.16&	\textbf{\underline{96.48}}&	89.85&	75.97&	87.06&	\underline{90.59} &	\underline{89.03} &	41.98&		80.99 \\
CDAN + EntMin & 93.60 &	72.10 &	82.02 &	\underline{62.33}&	89.82&	95.65&	\underline{91.51} &	77.32&	87.52 &	86.45 &	83.95 &	44.77 &	80.59 \\
CDAN + BNM & 94.80&	\underline{81.96}&	76.29&	54.60&	\underline{91.00}&	93.88&	86.90&	78.43&	88.99&	90.08&	86.88 &	44.46 &	80.69  \\
CDAN + LERM & 94.60 &	80.92 &	76.52 &	55.74&	90.36&	95.37&	88.37 &	\underline{80.28} &	\underline{89.45} &	89.83 &	83.36 &	\underline{52.97} &	\underline{81.48} \\
\midrule
SDAT  & 94.78&	83.79&	77.02&	66.10&	\textbf{\underline{93.57}} & 95.25 &	89.53&	80.99&	91.80&	90.13&	82.00&	54.02 & 83.25 \\
SDAT + EntMin & 94.51&	81.84&	\underline{79.45}&	\underline{68.78} &	92.84 &	96.29 &	\underline{89.84} &	\textbf{\underline{81.50}}&	\textbf{\underline{93.47}} & 85.71 &	80.88 &	51.87 &	83.08 \\
SDAT + BNM & \textbf{\underline{95.31}} &	81.41 &	76.78 &	62.22& 	93.35 &	93.98 &	87.22 &	78.75 &	89.98 &	\textbf{\underline{91.67}} &	84.87 &	52.20 	&	82.31 \\
SDAT + LERM & 94.94 &	\textbf{\underline{85.32}} &	77.72& 	63.54 &	92.85 &	\underline{96.41} &	89.38 &	79.05 &	91.36 &	91.30 &	\underline{85.55} & 	\textbf{\underline{55.58}} &	\textbf{\underline{83.58}} \\
\bottomrule
\vspace{-4ex}
\end{tabular}}
\end{table*}

\textbf{Evaluation on UDA Tasks}. We evaluate the LERM on three UDA benchmark datasets, \textit{i.e.}, Office-31 \citep{saenko2010adapting}, Office-Home \citep{venkateswara2017deep}, VisDA-2017 \citep{peng2017visda}, and ImageNet.
The Office-31 dataset contains three domains: Amazon (A), DSLR (D), and Webcam (W), with 4,110 images in 31 categories. 
We evaluate six transfer tasks built on the above domains.
The Office-Home dataset contains about 15,500 images from 65 categories within four domains: Art (Ar), Clipart (Cl), Product (Pr), and Real-World (Rw).
In those four domains, 12 transfer tasks are constructed for the evaluation.
The VisDA dataset contains $207,785$ images from 12 categories within two domains: Synthetic and Real.
Moreover, the transfer task on the ImageNet dataset is ImageNet$\rightarrow$ImageNet-Renditions (ImageNet-R) \citep{hendrycks2021many}. ImageNet-R contains 30,000 images of 200 categories in different formats. The same 200 categories on the ImageNet dataset are selected as the source domain. The textures and local image statistics of \mbox{ImageNet-R} are different from those of ImageNet images.
We combine the LERM with ERM, and state-of-the-art UDA methods such as CDAN~\citep{long2018conditional} and SDAT~\citep{rangwani2022closer} to learn from all the above tasks.
Additionally, we implement the EntMin and BNM in the same fashion as LERM. For each method, we run three random experiments and list the average classification accuracy. More experimental details are offered in \cref{a_UDA}.

The results on the Office-Home dataset for UDA are listed in \cref{officehome}. 
As can be seen, when combined with LERM, all the approaches achieve performance improvement. Specifically, the average accuracy of ERM+LERM outperforms ERM, ERM+EntMin, and ERM+BNM by 9.86\%, 3.43\%, and 0.47\%, respectively. 
Remarkably, ERM+LERM even performs better than well-established UDA methods such as DANN\citep{ganin2016domain}, AFN~\citep{xu2019larger}, and CDAN~\citep{long2018conditional}.
Note that the LERM does not explicitly match the distributions across domains. 
Instead, the LERM aligns the estimated label encodings built on unlabeled target samples to their corresponding ground-truth label encodings, 
while the ERM reduces the divergence between the predicted label encodings for the labeled source samples and their corresponding ground-truth label encodings. 
Thus, ERM+LERM implicitly reduces the distributional divergence between the source and target domains, leading to better transfer performance.
Those results testify to the superiority of LERM. 
The results on the Office-31 dataset are shown in \cref{office31}.
As can be seen, the LERM brings a substantial performance improvement of 7.93\% over ERM, and surpasses EntMin and BNM by a large margin of 4.27\% and 1.33\%, respectively.
Moreover, we can see that the performance of CDAN+LERM outperforms that of CDAN+EntMin and CDAN+BNM by 1.26\% and 0.55\%, respectively. 
Similar observations can be found in SDAT, highlighting the potential of the LERM. 
Table~\ref{tab:visda} reports the accuracies of each category and their average performance on the VisDA dataset. As can be seen, within each comparison group, the baseline approach equipped with LERM yields the best average performance. In particular, integrating LERM achieves an average performance improvement of 29.42\% over ERM. 
Moreover, as shown in \cref{ImageNet_UDA}, we verify the performance of LERM on the ImageNet$\rightarrow$ImageNet-R task. As can be seen, the integration of LERM enhances the performance of all methods.
In a nutshell, the LERM shows good transferability by combining existing UDA approaches or ERM.

\begin{table}[!t]
\footnotesize
\vspace{-6pt}
\centering
\caption{Accuracy (\%) comparison on the ImageNet$\rightarrow$ImageNet-R task under the UDA setting. The best performance of each task is marked in bold and the best performance in each comparison group is underlined.}
\label{ImageNet_UDA}
\setlength{\tabcolsep}{1.5mm}{
\begin{tabular}{cc}
\toprule 
Method & ImageNet$\rightarrow$ImageNet-R \\
\midrule
ERM & 35.57  \\
ERM + EntMin & 39.42  \\
ERM + BNM  & 39.75  \\
ERM + LERM & \underline{42.11}\\
\midrule
CDAN & 53.94 \\
CDAN + EntMin & 55.51 \\
CDAN + BNM & 55.80 \\
CDAN + LERM & \underline{56.80} \\
\midrule
SDAT & 55.47 \\
SDAT + EntMin & 56.48 \\
SDAT + BNM & 55.73 \\
SDAT + LERM & \underline{\textbf{57.62}} \\
\bottomrule
\vspace{-7ex}
\end{tabular}}
\end{table}

\textbf{Evaluation on SHDA Tasks}. We evaluate the LERM on the SHDA tasks involving text-to-text and text-to-image scenarios.
For the former, we adopt the Multilingual Reuters Collection dataset \citep{amini2009learning}.
As for the latter, we follow \citep{chen2016transfer,yao2019heterogeneous} to use the NUS-WIDE (N) \citep{chua2009nus} as the source domain and the ImageNet (I) dataset with three labeled samples per category as the target domain.
We combine the LERM with ERM, and state-of-the-art SHDA approach, \textit{i.e.}, KPG \citep{gu2022keypoint},
to handle all the above tasks. For a fair comparison, we implement EntMin and BNM in the same manner as LERM, and present the average classification accuracy of each method in three random experiments. More experimental details are given in \cref{a_HDA}.

\cref{text2text} lists the results of SHDA tasks. It can be observed that compared to EntMin and BNM, both ERM and KPG~\citep{gu2022keypoint} exhibit the most notable enhancements in performance when combined with LERM. In particular, the classification accuracy of ERM+LERM on N$\rightarrow$I is 79.04\%, which exceeds ERM, ERM+EntMin, and ERM+BNM by 11.62\%, 10.75\%, and 1.46\%, respectively. Overall, those results verify that even when handling heterogeneous samples, the LERM is still effective.

\begin{table*}[!t]\small
\centering
\caption{Accuracy (\%) comparison on the text-to-text and text-to-image datasets under the SHDA setting. The best performance of each task is marked in bold and the best performance in each comparison group is underlined. S5 and S10 indicate that there are five and ten labeled target samples in each category, respectively.}
\label{text2text}
\setlength{\tabcolsep}{1mm}{
\resizebox{1.7\columnwidth}{!}{
\begin{tabular}{cccccc|ccccc|c c}
\toprule Method & E$\rightarrow$S5 & F$\rightarrow$S5 & G$\rightarrow$S5 & I$\rightarrow$S5 & Average & E$\rightarrow$S10 & F$\rightarrow$S10 & G$\rightarrow$S10 & I$\rightarrow$S10 & Average & N$\rightarrow$I
 \\
\midrule
ERM & 60.94 & 61.03 & 60.18 & 61.97 & 61.03 & 68.70 & 68.54 & 68.63 & 69.06 & 68.73 & 67.42 \\
ERM + EntMin & 62.28 & 61.79 & 61.96 & 61.97 & 62.00 & 69.29 & 69.09 & 69.36 & 69.78 & 69.38 & 68.29  \\
ERM + BNM & 69.44 & 69.10 & \underline{69.58} & \underline{69.60} & 69.43 & 73.87 & 73.67 & 73.94 & 73.57 & 73.76 & 77.58  \\
ERM + LERM & \underline{\textbf{69.88}} & \underline{\textbf{70.24}} & 69.20 & 69.16 & \underline{69.62} & \underline{\textbf{74.56}} & \underline{\textbf{74.50}} & \underline{\textbf{74.21}} & \underline{74.24} & \underline{\textbf{74.38}} & \underline{79.04}  \\
\midrule 
KPG + ERM & 61.12 & 61.00 & 61.29 & 60.87 & 61.07 & 67.94 & 67.89 & 68.20 & 68.16 & 68.05 & 67.71  \\
KPG + BNM & 68.44 & 68.86 & 68.69 & 68.71 & 68.67 & 73.53 & 73.76 & 73.63 & 73.63 & 73.64 & 79.96  \\
KPG + LERM & \underline{69.20} & \underline{69.91} & \underline{\textbf{69.77}} & \underline{\textbf{69.98}} & \underline{\textbf{69.71}} & \underline{74.10} & \underline{74.22} & \underline{74.09} & \underline{\textbf{74.26}} & \underline{74.17} & \underline{\textbf{80.17}}  \\
\bottomrule
\end{tabular}}}
\vspace{-2ex}
\end{table*}

\subsection{Analysis}
\label{analysis}

\begin{figure}[t]
\centering
\setlength{\belowcaptionskip}{-10.cm}
\includegraphics[width=0.85\columnwidth]{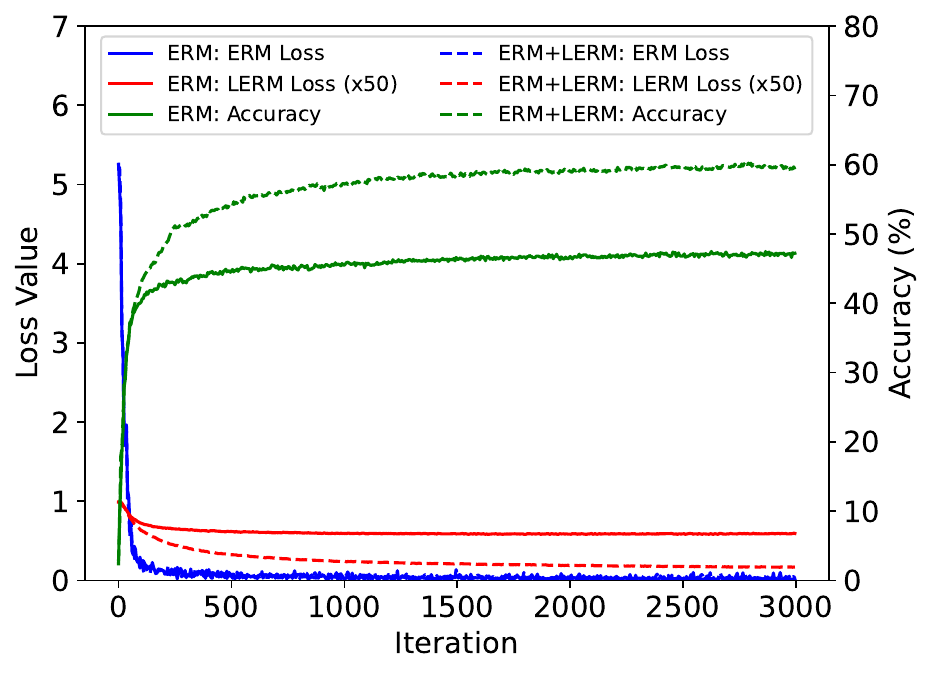}
\vspace{-3ex}
\caption{Comparison between LERM and ERM under the SSL setting.}
\label{fig:losscompare}
\vspace{-5ex}
\end{figure}

\textbf{Convergence}. 
We evaluate the convergence of LERM on the CIFAR-100 dataset for SSL tasks using four labeled samples per category. In \cref{fig:losscompare}, we plot the loss values of ERM and LERM within ERM and ERM+LERM, along with the testing accuracy curves for both methods. 
Several observations can be drawn from these results. (1) The loss values of ERM in both methods have experienced notable reductions, as they explicitly minimize the loss value of ERM in their objective functions. 
(2) When ERM is equipped with LERM, the accuracy curve improves by a large margin, which implies that LERM can effectively improve the performance of ERM. (3) The loss value of LERM in ERM + LERM is significantly lower than that of LERM in ERM, which is reasonable since ERM does not take LERM into account. Meanwhile, it is also one important reason for the previous observations. 
(4) The loss value of LERM in ERM + LERM first decreases gradually and then hardly changes as the number of iterations grows. Also, the accuracy of ERM + LERM first improves monotonically and then becomes stable with further iterations. Those trends collectively imply the convergence of LERM.

\textbf{Prediction Discriminability}. We perform experiments to analyze the prediction discriminability of LERM and EntMin for the SSL task built on the CIFAR-10 dataset with four labeled samples per category. Table \ref{tab:SSL_pp} shows the average entropy of predicted category distributions of unlabeled samples for the ERM, ERM+EntMin, and ERM+LERM methods on the CIFAR-10 dataset. We can observe that ERM+EntMin and ERM+LERM obtain much lower entropy values than ERM. Moreover, though ERM+LERM does not directly minimize the entropy over the predicted category distributions of unlabeled sample as ERM+EntMin did, it achieves a comparable entropy value with ERM+EntMin, which verifies properties (4) and (5) in \cref{theorem_m_property} and shows that both EntMin and LERM achieve good prediction discriminability as discussed in Section \ref{sec:comp_EntMin}.

\textbf{Prediction Diversity}. In \cref{AppendixPDA}, we conduct experiments to compare the prediction diversity of LERM and EntMin under class-imbalanced scenarios. The results indicate that, compared to EntMin, LERM can maintain prediction diversity even in category-imbalanced scenarios.

\textbf{Parameter Sensitivity and Feature Visualization}. We put the analysis of parameter sensitivity and feature visualization in \cref{PSFV}. 
Those results indicate that the LERM is not so sensitive to the trade-off hyperparameter $\lambda$ when its value is near the default setting presented in \cref{a_Parameter_Sensitivity}. 
Moreover, the t-SNE embeddings shown in Figure \ref{fig:tsne} demonstrate that LERM effectively aligns distributions across different domains, resulting in improved transfer performance.

\textbf{More Analysis Experiments}. Due to the page limit, we place additional analysis experiments in \cref{MoreAna}. These include an efficiency analysis of LERM, an examination of various losses within LERM, and a performance evaluation of LERM under the source-free domain adaptation setting~\cite{liang2020we}. The results demonstrate the effectiveness of LERM again.

\begin{table}[!t]\small
\centering
\vspace{-2ex}
\caption{Prediction discriminability comparison on the CIFAR-10 dataset under the SSL setting.}
\begin{tabular}{c|c}
\toprule Method & Entropy \\
     \midrule
     ERM & 0.3832  \\
     ERM + EntMin & 0.0266 \\
     ERM + LERM & 0.0440 \\ 
     \bottomrule
\end{tabular}
\label{tab:SSL_pp}
\vspace{-5ex}
\end{table}

\section{Conclusion}

In this paper, we propose the LERM to handle label insufficient scenarios, which can be regarded as an extension of the ERM to unlabeled samples. Similar to the ERM, the LERM adopts label encodings as guidance information to supervise the learning of unlabeled samples. However, different from the ERM, 
the LERM first estimates the label encodings by calculating the prediction means of unlabeled samples and then reduces the divergence between estimated and ground-truth label encodings. 
Thus, the prediction discriminability and diversity of unlabeled samples are guaranteed.
Theoretically, we analyze the properties of the prediction means for unlabeled samples, and the relationships between LERM and ERM, as well as between LERM and EntMin.
Experiments on several insufficient label scenarios validate the effectiveness of the LERM. 
Accordingly, we believe that the LERM has the potential to serve as an elegant and effective alternative to EntMin, thereby opening a new door for tackling unlabeled samples. Thus, applying the LERM to other label insufficient scenarios, for instance, open-set setting \cite{fang2021learning}, is our future research direction.

\section*{Acknowledgements}

This work is supported by NSFC key grant under grant no. 62136005, NSFC general grant under grant no. 62076118, and Shenzhen fundamental research program JCYJ20210324105000003.

\vspace{-1ex}

\section*{Impact Statement}

This paper presents work whose goal is to advance the field of Machine Learning. There are many potential societal consequences of our work, none of which we feel must be specifically highlighted here.

\nocite{langley00}

\bibliography{LERM_final}
\bibliographystyle{icml2024}

\newpage
\appendix
\onecolumn
\section{Theoretical Analyses}

\subsection{Proof for Theorem \ref{theorem_m_property}}
\label{prove_m_property}
(1) The sum of all elements in $\mathbf{m}_c^u$, \textit{i.e.}, $\mathbf{1}^T\mathbf{m}_c^u$, can be calculated as follows:
\begin{equation}
\mathbf{1}^T\mathbf{m}_c^u = \frac{\mathbf{1}^T \sum_{i = 1}^{n_u} (\widetilde{y}_{i,c}^u \widetilde{\mathbf{y}}_{i}^u)}{\sum_{i = 1}^{n_u} \widetilde{y}_{i,c}^u} = \frac{\sum_{i = 1}^{n_u} (\widetilde{y}_{i,c}^u(\mathbf{1}^T \widetilde{\mathbf{y}}_{i}^u))}{\sum_{i = 1}^{n_u} \widetilde{y}_{i,c}^u} = \frac{\sum_{i = 1}^{n_u} \widetilde{y}_{i,c}^u}{\sum_{i = 1}^{n_u} \widetilde{y}_{i,c}^u}=1.
\end{equation}

(2) Since $0 \le \widetilde{y}_{i,j}^u \le 1$, $\forall j \in \{1, \dots, C\}$, $\forall i \in \{1, \dots, n_u\}$, based on \eqref{m_c^u} and the Property (1) in \cref{theorem_m_property}, we have 
\begin{equation}
0 \le m_{c,j}^u \le 1, \forall j \in \{ 1, \dots, C \}.
\end{equation}

(3) Since $\widetilde{\mathbf{y}}_{i}^u$ is the ground-truth label encoding of sample $\mathbf{x}_i^u$,
$\widetilde{y}_{i,c}^u = 1$ if $\mathbf{x}_i^u$ belongs to category $c$, else $\widetilde{y}_{i,c}^u = 0$. Accordingly, we can calculate the $c$-th element of $\mathbf{m}_c^u$ as follows:
\begin{equation}
    m_{c, c}^u = \frac{\sum_{i = 1}^{n_u} {\widetilde{y}_{i,c}^{u}} \widetilde{y}_{i,c}^u}{ \sum_{i = 1}^{n_u} {\widetilde{y}_{i,c}^{u}} }
               = \frac{n_c^u}{n_c^u} = 1,
\end{equation}
where $n_c^u$ is the number of unlabeled samples belonging to category $c$. Similarly, the $k$-th ($\forall k \neq c$) element of $\mathbf{m}_c^u$ can be calculated as follows:
\begin{equation}
    m_{c, k}^u = \frac{\sum_{i = 1}^{n_u} {\widetilde{y}_{i,c}^{u}} \widetilde{y}_{i,k}^u}{\sum_{i = 1}^{n_u} {\widetilde{y}_{i,c}^{u}}}
               = \frac{0}{n_c^u} = 0, \forall k \neq c.
\end{equation}
Hence, $\mathbf{m}_c^u$ equals $\mathbf{e}_c$.

(4) Since $\mathbf{m}_c^u = \mathbf{e}_c$ for some $c \in \{1, \cdots, C\}$, we have $m_{c,c}^u=1$ and $m_{c,k}^u=0, \forall k\neq c$.

As $m_{c,k}^u = \frac{ \sum_{i = 1}^{n_u} \widetilde{y}_{i,c}^{u} \widetilde{y}_{i,k}^u }{ \sum_{i = 1}^{n_u} {\widetilde{y}_{i,c}^{u}} }$, we have 
\begin{equation}
\label{constraint2}
\sum_{i=1}^{n_u}\widetilde{y}_{i,c}^{u} \widetilde{y}_{i,k}^u = 0
\end{equation}
Since $\widetilde{y}_{i,c}^{u} \widetilde{y}_{i,k}^u\ge 0$, we have 
\begin{equation}
\widetilde{y}_{i,c}^{u} \widetilde{y}_{i,k}^u = 0,\  \forall k\ne c.\label{proof_4.1_cond}
\end{equation}

When $\widetilde{y}_{i,c}^{u}>0$, we have $\widetilde{y}_{i,k}^u = 0,\  \forall k\ne c$, which implies $\widetilde{y}_{i,c}^{u}=1$. 
When $\widetilde{y}_{i,c}^{u}$ equals 0, \eqref{proof_4.1_cond} naturally holds.

By combining the above two cases, \eqref{constraint2} is equivalent to that for some $c \in \{1, \cdots, C\}$ and any $i \in \{1, \cdots, n_u\}$, 
\begin{equation}
\label{proof_4.1_result1}
\widetilde{y}_{i,c}^u = 1, \widetilde{y}_{i,k}^u = 0, \forall k \neq c  \Rightarrow  \widetilde{\mathbf{y}}_{i}^u = \mathbf{e}_c 
\quad \textbf{or} \quad
\widetilde{y}_{i,c}^u = 0, 0 \leq \widetilde{y}_{i,k}^{u} \leq 1, \forall k \neq c.  
\end{equation}

Moreover, since $m_{c,c}^u = 1$, we obtain 
\begin{align}
\label{proof_4.2_eq1}
    m_{c,c}^u = \frac{ \sum_{i = 1}^{n_u} {\widetilde{y}_{i,c}^{u}} \widetilde{y}_{i,c}^u }{ \sum_{i = 1}^{n_u} {\widetilde{y}_{i,c}^{u}} } = 1 
     &\Rightarrow \sum_{i = 1}^{n_u} (({\widetilde{y}_{i,c}^{u}})^2 -{\widetilde{y}_{i,c}^{u}} )= 0
\end{align}
Since $0 \leq \widetilde{y}_{i,c}^{u} \leq 1$, $\forall i \in \{ 1, \cdots, n_u \}$, we can get $(\widetilde{y}_{i,c}^{u})^2\le \widetilde{y}_{i,c}^{u}$. To make \eqref{proof_4.2_eq1} hold, we must have $(\widetilde{y}_{i,c}^{u})^2= \widetilde{y}_{i,c}^{u}$, which implies that
\begin{equation}
\label{constraint1}
\widetilde{y}_{i,c}^{u} = 1 \quad \textbf{or} \quad \widetilde{y}_{i,c}^{u} = 0, \forall i \in \{ 1, \cdots, n_u \},
\end{equation}
which is consistent with \eqref{proof_4.1_result1}.

(5) Since $\mathbf{m}_c^u$ equals $\mathbf{e}_c$ for any $c \in \{1, \cdots, C\}$, based on \eqref{constraint1}, we have
\begin{equation}
\widetilde{y}_{i,c}^u = 1 \quad \textbf{or} \quad \widetilde{y}_{i,c}^u = 0, \forall i \in \{1, \cdots, n_u\}\  \forall c \in \{1, \cdots, C\}.
\end{equation}
Hence each element in $\widetilde{\mathbf{y}}_{i}^u$ is either 0 or 1, and because $\widetilde{\mathbf{y}}_{i}^u$ denotes the predicted probabilities for $C$ categories, it is evident that $\widetilde{\mathbf{y}}_{i}^u$ is a one-hot vector for any $i \in \{1, \cdots, n_u\}$.

\subsection{Proof for Theorem \ref{thm2}}
\label{appendixB}

Here we use the notations defined in Section \ref{sec:connect_LERM_ERM}.
Based on \eqref{R_emr}, the empirical risk for labeled samples is given by
\begin{equation}
\small
\begin{split}
\mathcal{R}_{emr}^l\!(f,g)& = \frac{1}{n_l} \!\!\sum_{c=1}^{C} \!\sum_{i=1}^{n_c^l} \mathcal{L}(\widetilde{\mathbf{y}}_{i}^{l,c}, \mathbf{e}_c) = \frac{1}{n_l} \sum_{c=1}^{C} \frac{n_c^l}{n_c^l} \sum_{i=1}^{n_c^l} \mathcal{L}(\widetilde{\mathbf{y}}_{i}^{l,c}, \mathbf{e}_c) 
\\ &
\ge \frac{1}{n_l} \sum_{c=1}^C n_c^l \mathcal{L}(\frac{1}{n_c^l} \sum_{i = 1}^{n_c^l} \widetilde{\mathbf{y}}_{i}^{l,c}, \mathbf{e}_c) = \frac{1}{n_l} \sum_{c=1}^C n_c^l \mathcal{L}(\mathbf{m}_{c}^{l}, \mathbf{e}_c) 
\\ &
\ge \frac{1}{C} \sum_{c=1}^C \mathcal{L} \big( \mathbf{m}_{c}^{l}, \mathbf{e}_c \big) = \mathcal{R}_{ler}^l(f,g),
\end{split}
\end{equation}
where $n_l$ denotes the total number of labeled samples, $n_c^l$ denotes the number of labeled samples in category $c$, and $C$ denotes the total number of categories.
The first inequality holds because of the convexity of the loss function, while the second inequality holds due to the new assumption $\frac{1}{n_l} \sum_{c=1}^{C} n_c^l \mathcal{L} (\mathbf{m}_{c}^{l}, \mathbf{e}_c) \ge \frac{1}{C}\sum_{c=1}^{C} \mathcal{L} (\mathbf{m}_{c}^{l}, \mathbf{e}_c)$.
Moreover, if $\mathcal{L} (\mathbf{m}_{c_1}^{l}, \mathbf{e}_{c_1})$ equals $\mathcal{L} (\mathbf{m}_{c_2}^{l}, \mathbf{e}_{c_2})$ for any two categories $c_1$ and $c_2$, the new assumption still holds.

\subsection{Proof for Theorem \ref{thm3}}
\label{appendixC}
Based on \eqref{EntMin}, the entropy for unlabeled samples, \textit{i.e.}, $R_{ent}(f,g)$, is given by
\begin{align}
\label{Ent_prove}
\mathcal{R}_{ent}(f,g) 
& = -\frac{1}{n_u} \sum_{i=1}^{n_u} (\widetilde{\mathbf{y}}_{i}^u)^\top \ln \widetilde{\mathbf{y}}_{i}^u \nonumber\\
& = -\frac{1}{n_u} \sum_{i=1}^{n_u} \sum_{c=1}^{C} \widetilde{y}_{i,c}^u \ln \widetilde{y}_{i,c}^u \nonumber\\
& = -\frac{1}{n_u} \sum_{c=1}^{C} \sum_{i=1}^{n_u} \widetilde{y}_{i,c}^u \ln \widetilde{y}_{i,c}^u \nonumber\\
& = -\frac{1}{n_u} \sum_{c=1}^{C} \sum_{j=1}^{n_u}\widetilde{y}_{j,c}^u 
\frac{1}{ \sum_{k=1}^{n_u} \widetilde{y}_{k,c}^u }
\sum_{i=1}^{n_u} \widetilde{y}_{i,c}^u \ln\widetilde{y}_{i,c}^u \nonumber\\
& = -\frac{1}{n_u} \sum_{c=1}^{C} \sum_{j=1}^{n_u}\widetilde{y}_{j,c}^u \sum_{i=1}^{n_u} \frac{\widetilde{y}_{i,c}^u}{ \sum_{k=1}^{n_u} \widetilde{y}_{k,c}^u } \ln\widetilde{y}_{i,c}^u \nonumber\\
& \ge -\frac{1}{n_u} \sum_{c=1}^{C} 
 \sum_{j=1}^{n_u} \widetilde{y}_{j,c}^u \ln \Big( \sum_{i=1}^{n_u} \frac{(\widetilde{y}_{i,c}^u)^2}{\sum_{k=1}^{n_u} \widetilde{y}_{k,c}^u} \Big) \nonumber\\
& = -\frac{1}{n_u} \sum_{c = 1}^{C} \sum_{j=1}^{n_u}\widetilde{y}_{j,c}^u \ln \Big(\frac{\sum_{i=1}^{n_u}(\widetilde{y}_{i,c}^u)^2}{\sum_{i=1}^{n_u}\widetilde{y}_{i,c}^u} \Big) \nonumber\\
& = -\frac{1}{n_u} \sum_{c = 1}^{C} \sum_{j=1}^{n_u}\widetilde{y}_{j,c}^u \ln \Big(\frac{\sum_{i=1}^{n_u} \widetilde{y}_{i,c}^u ( \mathbf{e}_c^\top \widetilde{\mathbf{y}}_{i}^u )}{\sum_{i=1}^{n_u}\widetilde{y}_{i,c}^u} \Big) \nonumber\\
& = -\frac{1}{n_u} \sum_{c = 1}^{C} \sum_{j=1}^{n_u}\widetilde{y}_{j,c}^u \mathbf{e}_c^\top \ln \Big(\frac{\sum_{i=1}^{n_u} \widetilde{y}_{i,c}^u \widetilde{\mathbf{y}}_{i}^u}{\sum_{i=1}^{n_u}\widetilde{y}_{i,c}^u} \Big) \\
& = -\frac{1}{n_u} \sum_{c = 1}^{C} \sum_{j=1}^{n_u}\widetilde{y}_{j,c}^u \mathbf{e}_c^\top \ln ( \mathbf{m}_c^u ) \nonumber\\
& \ge - \frac{1}{C} \sum_{c=1}^{C} \mathbf{e}_c^\top \ln(\mathbf{m}_c^u) \nonumber\\
& = \mathcal{R}_{ler}(f,g) ,\nonumber
\end{align}
where $n_u$ denotes the number of unlabeled samples. The first inequality holds because of the convexity of the negative logarithm function, \textit{i.e.}, $-\ln(\cdot)$, the second inequality holds due to the assumption $\frac{1}{n_u}\sum_{c=1}^{C}(\sum_{j=1}^{n_u}\widetilde{y}_{j,c}^u) \mathcal{L} (\mathbf{m}_c^u, \mathbf{e}_c) \ge \frac{1}{C}\sum_{c=1}^{C} \mathcal{L} (\mathbf{m}_c^u, \mathbf{e}_c)$, and \eqref{Ent_prove} holds because $\ln \big( a (\mathbf{e}_c^\top \widetilde{\mathbf{y}}_{i}^u) \big) = \ln ( a \widetilde{y}_{i, c}^u ) =  \mathbf{e}_c^\top \ln ( a \widetilde{\mathbf{y}}_{i}^u)$, where $a$ is a positive scalar.
It is worth mentioning that, if $\frac{\sum_{j=1}^{n_u} \widetilde{y}_{j,c}^u}{n_u} = \frac{1}{C}$, for all $c \in \{1, \cdots, C \}$, then $\frac{1}{n_u} \sum_{c=1}^{C} (\sum_{j=1}^{n_u} \widetilde{y}_{j,c}^u) \mathcal{L} (\mathbf{m}_c^u, \mathbf{e}_c) = \frac{1}{C}\sum_{c=1}^{C} \mathcal{L} (\mathbf{m}_c^u, \mathbf{e}_c)$.

\section{Implementation Details}
\label{ImpDetails}
The experiments on SSL and UDA tasks are conducted on a NVIDIA V100 GPU, and the experiments in SHDA tasks are conducted on a NVIDIA 3090 GPU.

\subsection{SSL}
\label{a_SSL}
The CIFAR-10 and CIFAR-100 datasets consist of 60,000 images with a resolution of 32$\times$32 pixels, categorized into 10 and 100 categories, respectively.
The DTD dataset contains 5,640 textural images in 47 categories.
The ImageNet-1K dataset consists of approximately one million images, distributed across 1,000 categories.

\begin{wraptable}{tr}{0.5\textwidth}
\small
\centering
\setlength\tabcolsep{2.2pt}
\vspace{-2ex}
\caption{Parameter $\lambda$ on SSL tasks.}
\label{hyperparameter}
\begin{tabular}{ccccc}
\toprule
Dataset               & CIFAR-10 & CIFAR-100 & DTD& ImageNet \\
\midrule
ERM + LERM                   & 0.1           & 50             & 1    &2000   \\
FlexMatch + LERM             & 1          & 50              & 50   & 100  \\
DST + LERM                   & 1          & 50              & 10   & 100  \\
\bottomrule
\end{tabular}
\end{wraptable}

The parameter $\lambda$ in \eqref{loss_ssl} for SSL tasks is shown in Table \ref{hyperparameter}, and the parameter $\mu$ in \eqref{loss_ssl} is set to 0.1.
We use mini-batch stochastic gradient descent (SGD) with a momentum of 0.9 as the optimizer, and the batch sizes of labeled and unlabeled samples are both set to 32 on the CIFAR-10, CIFAR-100, and DTD datasets and 512 on the ImageNet dataset.
Following \citep{chen2022debiased}, we use random-resize-crop and RandAugment \citep{cubuk2020randaugment} for strong augmentation and random-horizontal-filp for weak augmentation.
To ensure a fair comparison, all methods utilize the same backbone for each dataset. 
Specifically, we utilize ResNet-18 \citep{he2016deep} for the CIFAR-10 dataset, while ResNet-50 \citep{he2016deep} is adopted on the CIFAR-100, DTD, and ImageNet datasets. Those backbone architectures are pretrained on the ImageNet dataset \citep{deng2009imagenet}, except for tasks on the ImageNet dataset, for which the ResNet-50 backbone is trained from scratch.

\begin{wraptable}{tr}{0.5\textwidth}
\small
\centering
\setlength\tabcolsep{2.2pt}
\vspace{-6ex}
\caption{Parameter $\lambda$ on UDA tasks.}
\label{hyperparameter_UDA}
\begin{tabular}{ccccc}
\toprule
Dataset               & Office-31 & Office-Home & VisDA  & ImageNet \\
\midrule
ERM + LERM              & 10        & 10     & 10  & 100 \\
CDAN + LERM             & 0.1       & 10     & 1   &  100   \\
SDAT + LERM             & 1          & 10     & 0.01   &  50 \\
\bottomrule
\end{tabular}
\vspace{-4ex}
\end{wraptable}

\subsection{UDA}
\label{a_UDA}
For the UDA tasks, $\lambda$ in \eqref{loss_uda} is shown in Table \ref{hyperparameter_UDA}.
We set the batch sizes of both domains to 32. The optimizer is a mini-batch SGD method with a momentum of 0.9 and a learning rate annealing strategy in \citep{ganin2016domain}.
For a fair comparison, we use ResNet-101 pretrained on ImageNet as the backbone for tasks on the VisDA dataset, and ResNet-50 pretrained on ImageNet as the backbone for tasks on other datasets.

\subsection{SHDA}
\label{a_HDA}
In the text-to-text scenario, the Multilingual Reuters Collection dataset consists of about 11,000 articles from six categories written in English (E), French (F), German (G), Italian (I), and Spanish (S). Following \citep{hsieh2016recognizing,li2013learning,yao2019heterogeneous}, we treat S as the target domain and the remaining datasets are used as the source domains. Also, we randomly pick up 100 labeled articles per category as source samples, while in the target domain, there are randomly selected $l$ labeled samples (\textit{i.e.}, $l = 5, 10$) and 500 unlabeled samples in each category. 
The reduced dimensions of E, F, G, I, and S are 1,131, 1,230, 1,417, 1,041, and 807, respectively. 
In the text-to-image scenario, eight shared categories from both domains are chosen. For the source domain, we randomly select 100 texts per category as the labeled samples. As for the target domain, we randomly single out three images from each category as the labeled samples, and the remaining images are considered as the unlabeled samples. 
We extract the 64-dimensional features from the fourth hidden layer of a five-layer neural network as the text features, and the 4096-dimensional $DeCAF_6$ features \citep{donahue2014decaf} are extracted to represent the images in the target domain. 

In addition, for the SHDA tasks, we implement $g_s(\cdot)$ and $g_t(\cdot)$ in \eqref{loss_shda} by using a one-layer fully connected network with the Leaky ReLU \citep{maas2013rectifier} activation function, respectively. Analogously, we adopt a one-layer fully connected network with the softmax activation function for $f(\cdot)$ in \eqref{loss_shda}. In addition, following \citep{yao2019heterogeneous}, we utilize full-batch gradient descent with the Adam optimizer \citep{kingma2014adam} to optimize $\{ f(\cdot), g_s(\cdot), g_t(\cdot) \}$, and the learning rate is set to 0.001. We maintain a fixed value of $\lambda=1$ and $\tau=0.01$ in \eqref{loss_shda} for all the experiments.

\section{Prediction Diversity Analysis}
\label{AppendixPDA}

\begin{figure}[t]
\centering
\subfigure[Ground-truth category distribution]{
    \includegraphics[width=0.3\textwidth]{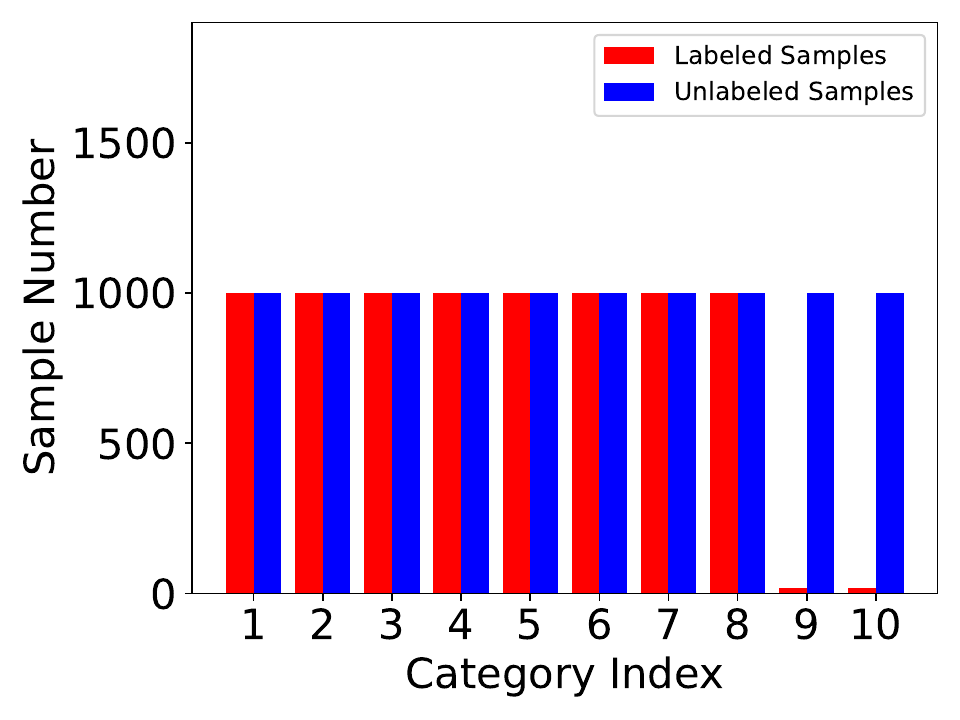}
    \label{fig:prior}
}
\subfigure[Predicted category distribution by ERM + EntMin]{
    \includegraphics[width=0.3\textwidth]{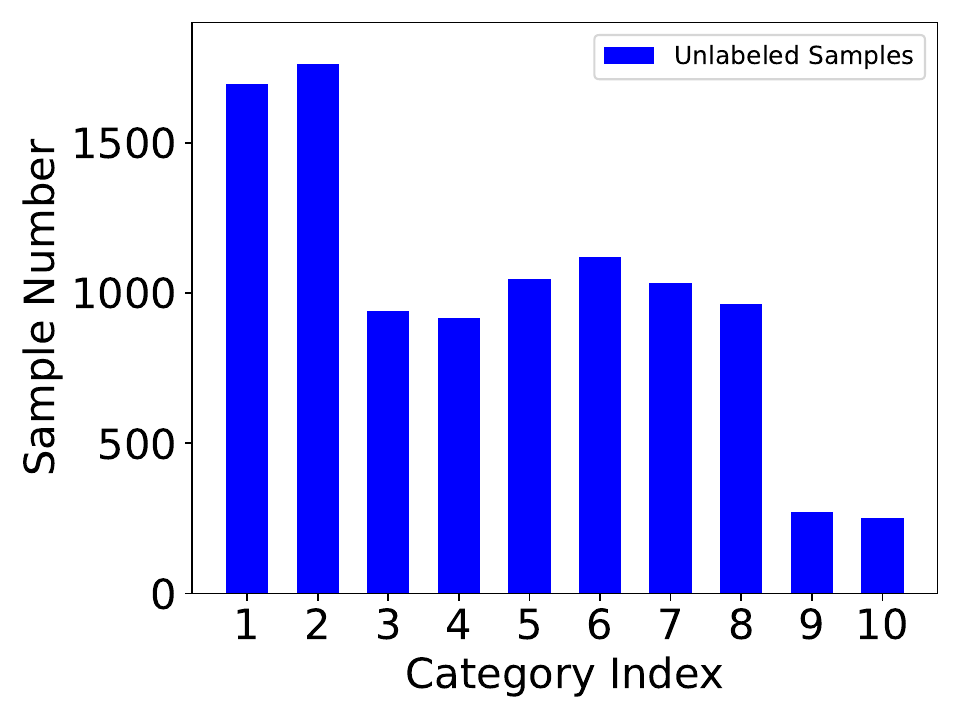}
    \label{fig:entmin}
}
\subfigure[Predicted category distribution by ERM + LERM]{
    \includegraphics[width=0.3\textwidth]{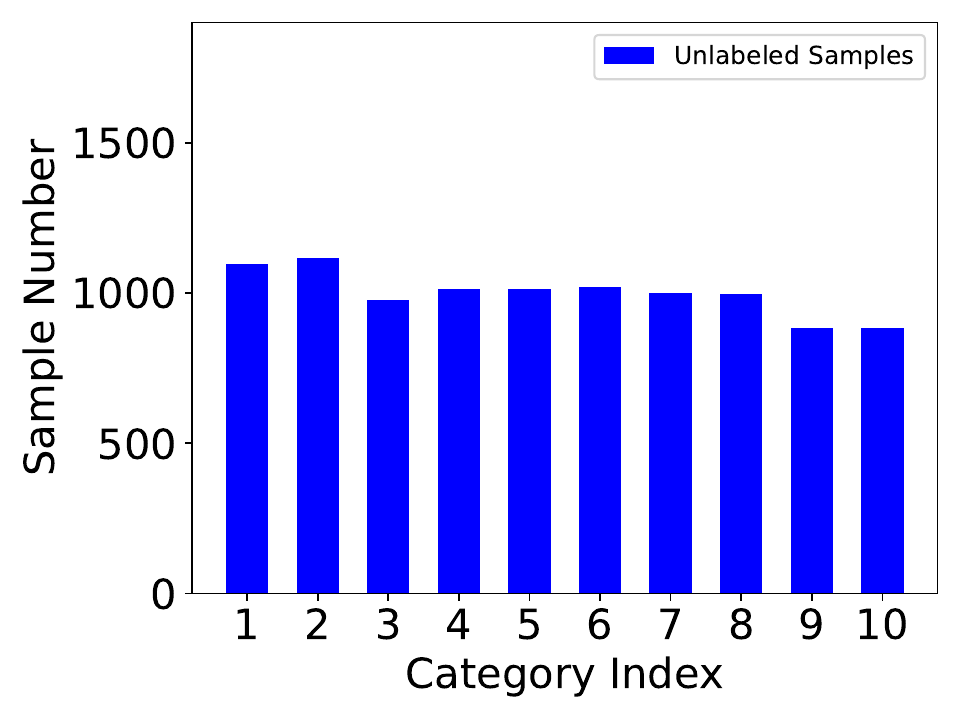}
    \label{fig:crm}
}
\vskip -.1in
\caption{Empirical evaluation of prediction diversity on the SSL task on CIFAR-10 dataset under the class-imbalanced setting. (a) The ground-truth category distributions of the labeled and unlabeled samples. (b) The predicted category distribution of the unlabeled samples by ERM + EntMin. (c) The predicted category distribution of the unlabeled samples by ERM + LERM.}
\label{fig:diversity_ssl}
\vspace{-3ex}
\end{figure}

We conduct experiments to compare the prediction diversity of LERM and EntMin under class-imbalanced scenarios. 
To this end, we rebuild the SSL task on the CIFAR-10 dataset into a category-imbalanced setting.
Specifically, as shown in Figure~\ref{fig:prior}, we randomly choose 1,000 labeled samples from each category for the first eight categories.
As for the last two categories, we randomly pick 20 labeled samples per category. 
Also, 1,000 unlabeled samples per category are used for testing. The experimental results are plotted in Figure~\ref{fig:diversity_ssl}. As can be seen, compared with ERM + EntMin, ERM + LERM is less susceptible to the impact of category imbalance. 
Moreover, the average classification accuracies of ERM + EntMin and ERM + LERM are $79.47\%$ and $92.97\%$, respectively. And the average F1 scores of ERM + EntMin and ERM + LERM are $77.09$ and $93.27$, respectively. 
Those results indicate that the LERM can effectively preserve prediction diversity even in category-imbalanced scenarios.

\begin{wrapfigure}{t}{0.3\textwidth}
  \centering
    \includegraphics[width=0.3\textwidth]{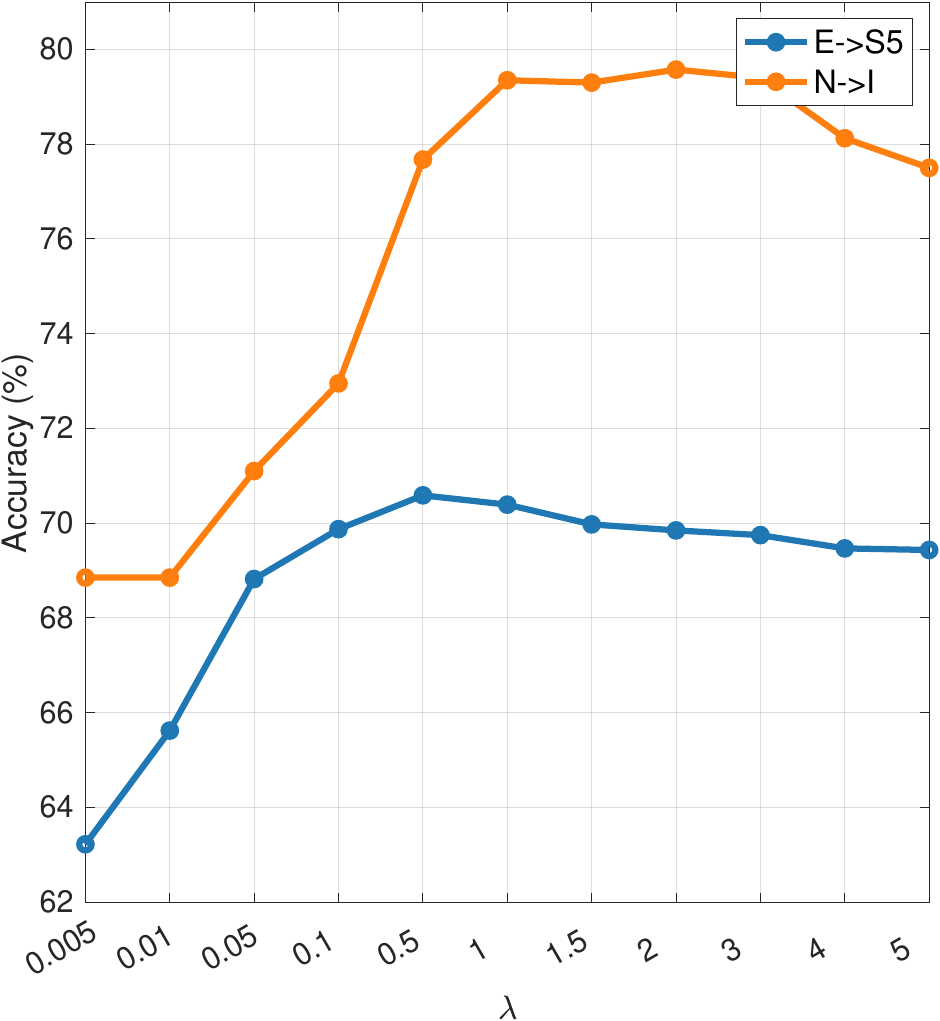}
    \vspace{-8mm}
    \caption{Parameter sensitivity analysis on the SHDA tasks of E$\rightarrow$S5 and N$\rightarrow$I.}
    \label{fig:sensitive}
\vspace{-1ex}
\end{wrapfigure}

\section{Parameter Sensitivity and Feature Visualization}
\label{PSFV}

\subsection{Parameter Sensitivity}
\label{a_Parameter_Sensitivity}
Note that the only hyperparameter introduced by our method is $\lambda$.
We investigate the parameter sensitivity on the SHDA tasks of E$\rightarrow$S5 and N$\rightarrow$I. We mainly investigate the sensitivity of the parameter $\lambda$ in \eqref{loss_shda}, as it controls the importance of the LERM. 
Note that when $\lambda = 0$, ERM + LERM degenerates to the original ERM. According to the accuracy \textit{w.r.t.} ${\lambda}$ shown in Figure \ref{fig:sensitive}, in the initial phase of increasing the value of $\lambda$, the significant performance improvement indicates the effectiveness of LERM. 
As $\lambda$ increases further, the performance improves and then saturates around $\lambda = 1$. After that, the performance decreases gradually with the increase in the value of $\lambda$. One possible reason is that, with a large $\lambda$, the model excessively focuses on LERM with unlabeled samples and neglects ERM on labeled samples. Those results indicate that the LERM is not so sensitive to $\lambda$ when its value is near the default setting, \textit{i.e.}, $\lambda = 1$.
In summary, the default setting, \textit{i.e.}, $\lambda = 1$, can lead to good performance on both tasks, which suggests that this setting is a good choice for $\lambda$.

Moreover, we further conduct experiments to analyze the sensitivity of $\lambda$ in ERM+LERM on the CIFAR-100 dataset under the semi-supervised learning setting and the Office-31 dataset under the unsupervised domain adaptation setting, respectively. Specifically, we experiment with a wide range for $\lambda$ and present the results in Tables~\ref{tab:SSL_sensitive} and~\ref{tab:office31_sensitive}.
According to the results under the SSL and UDA settings, we can see that as $\lambda$ increases, the performance of ERM + LERM initially shows a gradual improvement, followed by a slight degradation.
On the other hand, ERM + LERM performs well on all the UDA tasks by taking the default parameter value, \textit{i.e.}, $\lambda = 10$, which indicates $\lambda$ is not so sensitive across various UDA tasks when $\lambda$ equals 10. All the results once again indicate the LERM is not so sensitive to $\lambda$ when its value is close to the default setting, \textit{i.e.,} $\lambda = 50$ for SSL and $\lambda = 10$ for UDA. Furthermore, it is worth noting that ERM + LERM with all positive values of $\lambda$ outperforms ERM (\textit{i.e.}, $\lambda=0$), showing the effectiveness of our LERM method.

\begin{table}[t]
    \begin{minipage}[t]{0.4\textwidth}
        \footnotesize
\centering
\caption{Parameter sensitivity analysis on the CIFAR-100 dataset under the SSL setting. The best performance of each task is marked in bold.}
\label{tab:SSL_sensitive}
\begin{tabular}{c|cc|c}
\toprule Dataset               & \multicolumn{3}{c}{CIFAR-100}   \\
\cmidrule(r){2-4}  
\# Label per category & \multicolumn{2}{c|}{1} & 4     \\
$\lambda$& Top-1     & Top-5 & Top-1    \\
\midrule
0   & 23.58 & 47.51  & 47.18  \\
0.01   & 23.62 & 48.19  & 48.12  \\
0.1  & 23.53  & 47.92 & 47.96  \\
1   & 24.88    & 50.91 & 48.98  \\
5   & 29.13  & 57.98 & 53.20  \\
10 & \textbf{31.84} & 61.24 & 56.23 \\
50 & 30.15  & \textbf{61.33} & \textbf{60.19} \\
    100 & 29.26 & \textbf{61.33} & 58.83 \\
    \bottomrule
    \end{tabular}
    \end{minipage}
    \hspace{2px}
    \begin{minipage}[t]{0.55\textwidth}
        \centering
\caption{Parameter sensitivity analysis on the Office-31 dataset under the UDA setting. The best performance of each task is marked in bold.}
\label{tab:office31_sensitive}
\vspace{10px}
\setlength{\tabcolsep}{1.5mm}{
\begin{tabular}{ccccccc @{\hskip 0.1in} c}
\toprule $\lambda$ & A$\rightarrow$D        & A$\rightarrow$W  & D$\rightarrow$W        & W$\rightarrow$D    & D$\rightarrow$A    & W$\rightarrow$A        & Average \\
\midrule
0  & 81.15 & 77.00 & 96.60 & 99.00 & 63.98 & 64.01 & 80.29 \\
0.01  & 83.53 &	83.40 &	97.86 &	99.60 &	62.34 &	62.30 &	81.51  \\
0.1  & 83.94  &	85.03  &	98.24  &	99.60  &63.33 & 62.62 &	82.13  \\
1 & 88.96 & 91.70 & 98.62 & \textbf{100.00} &	69.33 &	69.76 &	86.40 \\
5 & 90.96 & 92.45 & \textbf{98.74} & \textbf{100.00} & 71.99 & 72.67 &	87.80 \\
10 & \textbf{92.37} &	\textbf{92.96} &	\textbf{98.74} &	\textbf{100.00} &	\textbf{72.45} 	&\textbf{72.81} &	\textbf{88.22}  \\
50 & 89.36 & 	86.42 & 	97.61 & 	98.39 &	68.34 &	67.31 &	84.57 \\
100 & 89.76 &	83.02 &	97.23 &	97.99 	&64.86 &	64.50& 	82.89  \\
\bottomrule
\end{tabular}}
    \end{minipage}
\end{table}

\subsection{Feature Visualization}
\label{a_Feature_Visualization}
\begin{figure}[t]
  \centering
    \subfigure[ERM]{\includegraphics[width=0.3\textwidth]{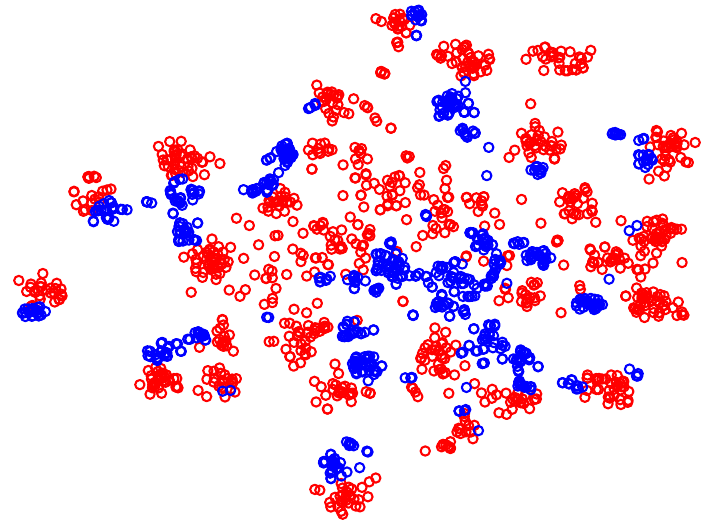}
\label{fig:tsne_erm}}
\hspace{10px}
\subfigure[ERM + LERM]{\includegraphics[width=0.3\textwidth]{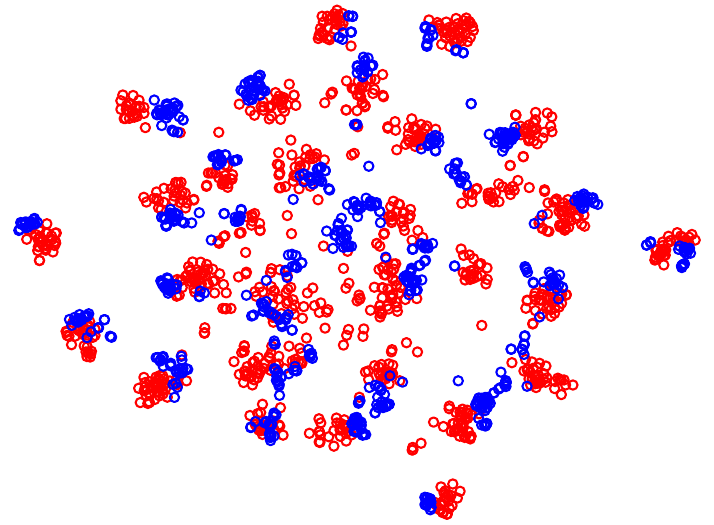}
\label{fig:tsne_ccl}}
    \caption{t-SNE visualization for the UDA task A$\rightarrow$D on the Office-31 dataset. The red and blue circles represent the source and target features, respectively.}
    \label{fig:tsne}
    \vspace{-5mm}
\end{figure}

Figure \ref{fig:tsne} visualizes the t-SNE embeddings \citep{van2008visualizing} of the learned source and target features on the UDA task of A$\rightarrow$D for ERM and ERM + LERM. 
As can be seen, compared with ERM alone, ERM + LERM can better align the distributions across domains. 
The primary reason is that LERM aligns the estimated label encodings of unlabeled target samples with ground-truth label encodings. 
Also, ERM minimizes the discrepancy across the predicted label encodings of labeled source samples and their corresponding ground-truth label encodings. 
As a result, ERM + LERM implicitly aligns the distributions of the source and target domains, thereby producing a positive transfer.

\begin{table}[!tbph]
\footnotesize
\centering
\caption{Average time cost (in seconds) per training epoch on the ImageNet dataset under the SSL setting.}
\begin{tabular}{c|c}
\toprule Method & Time (s) \\
     \midrule
     ERM & 492.92  \\
     ERM + EntMin & 947.57  \\
     ERM + BNM & 967.20 \\
     ERM + LERM & 944.45 \\
         \midrule
     FlexMatch & 871.69 \\
     FlexMatch + EntMin &918.88 \\
     FlexMatch + BNM & 930.23 \\
     FlexMatch + LERM & 876.53 \\
         \midrule
     DST & 941.06 \\
     DST + EntMin & 965.72  \\
     DST + BNM & 976.76\\
     DST + LERM & 953.78 \\
     \bottomrule
\end{tabular}
\label{tab:time}
\end{table}

\section{More Analysis Experiments}
\label{MoreAna}

\subsection{Computational Complexity}
In Table~\ref{tab:time}, we record the average time cost (in seconds) per training epoch over five epochs on the ImageNet dataset under the semi-supervised learning setting. The observations are summarized as follows: (i) ERM is faster than other methods since it only utilizes labeled samples for learning; (ii) When integrating LERM into ERM, LERM is slightly faster than both EntMin and BNM, which implies that LERM is more computationally efficient; and (iii) When integrating LERM into some semi-supervised methods (\textit{e.g.}, FlexMatch and DST), it imposes a relatively small computational burden when compared with baselines. Overall, all the results demonstrate the efficiency of LERM.

\begin{wraptable}{tr}{0.5\textwidth}
\centering
\vspace{-4.5ex}
\caption{Accuracy (\%) comparison of different losses in LERM on the CIFAR-10 dataset for SSL. The best performance of each task is marked in bold.}
\label{table:loss}
\resizebox{0.4\columnwidth}{!}{
\begin{tabular}{c|cc|c}
\toprule Dataset            & \multicolumn{3}{c}{CIFAR-10}    \\
\cmidrule(r){2-4}  
\# Label per category & \multicolumn{2}{c|}{1} & 4     \\
& Top-1      & Top-5    & Top-1  \\
\midrule
\textit{KL} loss     & 35.23      & 79.12   & 71.24 \\
$L_2$ loss  & 37.08      & 80.77    & 74.40   \\
$L_1$ loss  & \textbf{38.22}      & \textbf{80.82}  & \textbf{75.57}  \\
\bottomrule
\end{tabular}}
\end{wraptable}

\subsection{Loss for LERM}
\label{appendix:loss}
We provide a comparative analysis of the accuracy achieved by various losses in the LERM framework on the CIFAR-10 dataset for SSL. As can be seen in Table \ref{table:loss}, 
LERM with the $L_1$ loss or $L_2$ loss is better than LERM with \textit{KL} loss, which implies that \textit{KL} loss may not be suitable for reducing the divergence between the estimated and ground-truth label encodings. One possible reason is that for each estimated label encoding, the \textit{KL} loss only forces the probability in the position corresponding to the ground-truth label to become one, but does not constrain the probabilities in other positions, whereas the $L_1$ and $L_2$ losses do. Moreover, LERM with the $L_1$ loss slightly outperforms LERM with the $L_2$ loss.
One possible reason is that the $L_1$ loss is more robust than the $L_2$ loss as the $L_2$ loss is easier to be affected by large values. Consequently, we empirically choose the $L_1$ loss in the implementation of LERM.

\subsection{Evaluation on Source-Free Domain Adaptation Tasks}

Source-Free Domain Adaptation (SFDA)~\cite{liang2020we} aims to adapt a well-trained model from a source domain to a related target domain, without requiring access to the source samples during adaptation. 
To apply the LERM to the SFDA setting, we finetune the pretrained source model by minimizing the label-encoding risk on unlabeled target samples. We perform experiments on the VisDA dataset for SFDA. The results are presented in Table~\ref{tab:sourcefree_visda}. Here, the ``Source-only'' method refers to testing the pretrained source model in the target domain directly. As can be seen, LERM is still effective under the SFDA setting, outperforming the source-only method by a large margin. One reason is that LERM is based on a pretrained source model, which provides a reasonable and non-random initial prediction for unlabeled target samples. Thus, those results verify the effectiveness of LERM again. Also, we think that LERM may be a promising method for SFDA.

\begin{table*}[!htbp]\small
\vspace{-3ex}
\centering
\caption{Accuracy (\%) comparison on the VisDA dataset under the SFDA setting.  The best performance of each task is marked in bold.}
\label{tab:sourcefree_visda}
\setlength{\tabcolsep}{0.5mm}{
\begin{tabular}{*{14}{c}}
\toprule Method & aeroplane & bicycle & bus & car & horse & knife &	motor &	person	& plant & skate & train & truck & Average \\
\midrule
Source-only  & 54.03 & 17.61 & 49.08 & 75.67 & 60.65 & 4.34 & 84.54 & 20.28 & 69.69 & 31.70 & 80.81 & 7.91 & 46.36 \\
LERM & \textbf{92.92} & \textbf{73.04} & \textbf{80.32} & \textbf{60.19} & \textbf{89.38} & \textbf{94.80} & \textbf{84.70} & \textbf{78.15} & \textbf{79.42} & \textbf{82.16} & \textbf{82.93} & \textbf{52.97} & \textbf{79.25} \\
\bottomrule
\end{tabular}}
\end{table*}


\end{document}